\documentclass[10pt,journal,compsoc]{IEEEtran}

\ifCLASSOPTIONcompsoc

\usepackage{times}
\usepackage{epsfig}
\usepackage{graphicx}
\usepackage{amsmath}
\usepackage{amssymb}
\usepackage{color}
\usepackage{algorithm}
\usepackage{subfigure}
\usepackage{epstopdf}
\usepackage{booktabs}
\usepackage{array}
\usepackage{multirow}
\usepackage{hhline}
\usepackage{makecell}
\usepackage{framed}
\usepackage{enumitem}
\usepackage{algpseudocode}
\usepackage{ulem}
\usepackage[misc]{ifsym}

\newcommand{\Tref}[1]{Table~\ref{#1}}

\newcommand{\Fref}[1]{Figure~\ref{#1}}
\newcommand{\Sref}[1]{Section~\ref{#1}}

\newcommand{\fref}[1]{Fig.~\ref{#1}}

\newcommand{\etal}{\textit{et al}. }

\graphicspath{ {images/} }

\begin{document}
	
\title{Deep Model Intellectual Property Protection via Deep Watermarking}
	
	\author{Jie Zhang, Dongdong Chen,
		Jing Liao, Weiming Zhang, Huamin Feng,\\
		Gang Hua,~\IEEEmembership{Fellow,~IEEE}, and Nenghai Yu
		
		\IEEEcompsocitemizethanks{
			\IEEEcompsocthanksitem Jie Zhang, Weiming Zhang and Nenghai Yu are with School of Cyber Science and  Security, University of Science and Technology of China, Hefei, Anhui 230026, China.\protect\\
			E-mail: \{zjzac@mail., zhangwm@, ynh@\}ustc.edu.cn
			\IEEEcompsocthanksitem Dongdong Chen is with Microsoft Research, Redmond, Washington 98052, USA.\protect\\
			E-mail: cddlyf@gmail.com
			\IEEEcompsocthanksitem Jing Liao is with Department of Computer Science, City University of Hong Kong \protect\\
			E-mail: jingliao@cityu.edu.hk
			\IEEEcompsocthanksitem Huaming Feng is with Beijing Electronic Science and Technology Institute \protect\\
			E-mail: fenghm@besti.edu.cn
			\IEEEcompsocthanksitem Gang Hua is with Wormpex AI Research LLC, WA 98004, US.\protect\\
			E-mail: ganghua@gmail.com
		}
	}

	\markboth{Journal of \LaTeX\ Class Files,~Vol.~14, No.~8, August~2015}%
	{Shell \MakeLowercase{\textit{et al.}}: Bare Demo of IEEEtran.cls for Computer Society Journals}
	
	\IEEEtitleabstractindextext{%
		\begin{abstract}
	Despite the tremendous success, deep neural networks are exposed to serious IP infringement risks. Given a target deep model, if the attacker knows its full information, it can be easily stolen by fine-tuning. Even if only its output is accessible, a surrogate model can be trained through student-teacher learning by generating many input-output training pairs. Therefore, deep model IP protection is important and necessary. However, it is still seriously under-researched. In this work, we propose a new model watermarking framework for protecting deep networks trained for low-level computer vision or image processing tasks. Specifically, a special task-agnostic barrier is added after the target model, which embeds a unified and invisible watermark into its outputs. When the attacker trains one surrogate model by using the input-output pairs of the barrier target model, the hidden watermark will be learned and extracted afterwards. To enable watermarks from binary bits to high-resolution images, a deep invisible watermarking mechanism is designed. By jointly training the target model and watermark embedding, the extra barrier can even be absorbed into the target model. Through extensive experiments, we  demonstrate the robustness of the proposed framework, which can resist attacks with different network structures and objective functions.
		\end{abstract}
		\begin{IEEEkeywords}
			Deep Model IP Protection, Model Watermarking, Image Processing
		\end{IEEEkeywords}
	}
	
	\maketitle
	
	\IEEEdisplaynontitleabstractindextext
	
	\IEEEpeerreviewmaketitle
	
	\IEEEraisesectionheading{
		\section{Introduction}\label{sec:introduction}
	}

	\IEEEPARstart{I}{n} recent years, deep learning has revolutionized a wide variety of artificial intelligence fields such as image recognition \cite{krizhevsky2012imagenet,he2016deep}, medical image processing \cite{hong2017encase,hong2019combining,zhang2017rebuild,hong2017event2vec}, speech recognition \cite{graves2013speech,zhang2017towards} and natural language processing \cite{vaswani2017attention}, and outperforms traditional state-of-the-art methods by a large margin. Despite its success, training a good deep model is not a trivial task and often requires the dedicated design of network structures and learning strategies, a large scale high-quality labeled data and massive amount of computation resources, all of which are expensive and full of great business value. For some companies, these trained models are indeed their core competitiveness. Therefore, IP protection for deep models is not only important but also essential for their success. 
	
    However, compared to media IP protection, deep model IP protection is much more challenging because of the powerful learning capacity of deep models \cite{uchida2017embedding,adi2018turning}. In other words, given one specific task, numerous structures and weight combinations can obtain similar performance. This actually makes the IP infringement rather convenient. In the white-box case, where the full information including the detailed network structure and weights of the target model is known, one typical and effective attack way would be fine-tuning or pruning based on the target model on new datasets. Even in the black-box case where only the output of the target model can be accessed, the target model can still be stolen by training another surrogate model to imitate its behavior through student-teacher learning \cite{NIPS20145484,hinton2015distilling}. Specifically, we can first generate a large scale of input-output training pairs based on the target model, and then directly train the surrogate model in a supervised manner by regarding the outputs of the target model as ground-truth labels.

   Very recently, some preliminary research works \cite{uchida2017embedding,adi2018turning,zhang2018protecting,merrer2017adversarial} emerge for deep model IP protection. They often either add a weight regularizer into the loss function to make the learned weight have some special patterns or use the predictions of a special set of indicator images as the watermarks. Though these methods work pretty well to some extent, they only consider the classification task and the white-box attacks like fine-tuning or pruning. But in real scenarios, labeling the training data for low-level computer vision or image processing tasks is much more complex and expensive than classification tasks, because their ground-truth labels should be pixel-wise precise. Examples include removing all the ribs in Chest X-ray images and the rain streaks in real rainy images. In this sense, protecting such image processing models is more valuable. Moreover, because the original raw model does not need to be provided in most application scenarios, it can be easily encrypted with traditional algorithms to resist the white-box attack. Therefore, more attention should be paid to the black-box surrogate model attack.  

   Motivated by this, this paper studies the IP protection problem for image processing networks that aims to resist the challenging surrogate model attack \cite{zhang2020model}. Specifically, a new model watermarking framework is proposed to add watermarks into the target model. When the attackers use a surrogate model to imitate the behavior of the watermarked model, the designed watermarking mechanism should be able to extract pre-defined watermarks out from outputs of the learned surrogate model.

\begin{figure}
	    \centering
	    \includegraphics[width=\linewidth]{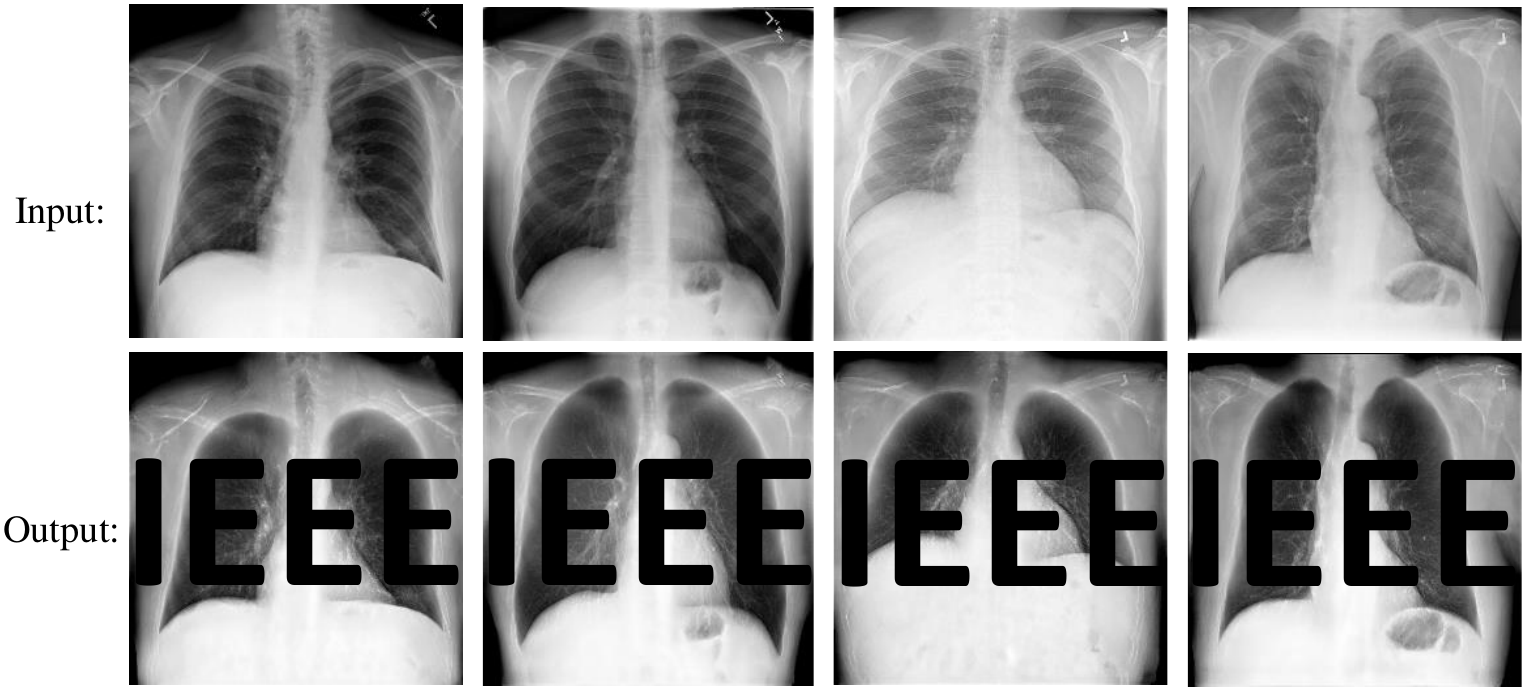}
	    \caption{The simplest watermarking mechanism by adding unified visible watermarks onto output of the target model, which will sacrifice the visual quality and usability.}
	    \label{fig:simple_vis_wm}
	\end{figure}

For better understanding, before diving into the model watermarking details, let us first discuss the simplest spatial visible watermarking mechanism shown in \Fref{fig:simple_vis_wm}. Suppose that we have a lot of input-output training pairs and we manually add a unified visible watermark template to all the outputs. Intuitively, if a surrogate model is trained on such pairs with the simple L2 loss, the learned model will learn this visible watermark into its output to get lower loss. That is to say, given one target model, if we forcibly add one unified visible watermark into all its output, it can resist the plagiarism from other surrogate models to some extent. However, the biggest limitation of this method is that the added visible watermarks will seriously degrade the visual quality and usability of the target model. Another potential threat is that attackers may use image editing tools like Photoshops to manually remove all the visible watermarks.

Inspired by the above observation, we propose a general deep invisible model watermarking framework as shown in \Fref{fig:framework}. Given a target model $\mathbf{M}$ to be protected, we denote its original input and output images as domain $\mathbf{A}$  and $\mathbf{B}$ respectively. Then a spatial invisible watermark embedder $\mathbf{H}$ is used to hide a unified target watermark ${\delta}$ into all the output images in the domain $\mathbf{B}$ and generate a new domain $\mathbf{B'}$. Different from the aforementioned simple visible watermarks, all the images in the domain $\mathbf{B'}$ should be visually consistent to domain $\mathbf{B}$. Symmetrically, given the images in domain $\mathbf{B'}$, another watermark extractor $\mathbf{R}$ will extract the watermark ${\delta'}$ out, which is consistent to ${\delta}$. To protect the IP of $\mathbf{M}$, we will pack $\mathbf{H}$ and $\mathbf{M}$ into a whole for deployment. The key hypothesis here is that when the attacker uses $\mathbf{A}$ and $\mathbf{B'}$ to learn a surrogate model $\mathbf{SM}$, $\mathbf{R}$ can still extract the target watermark out from the output $\mathbf{B''}$ of $\mathbf{SM}$.

	We first test the effectiveness of our framework by using traditional spatial invisible watermarking algorithms such as \cite{kutter1999watermarking,voloshynovskiy2000content} and \cite{voloshynovskiy2001multibit}, which work well for some special surrogate models but fail for most other ones. Another common limitation is that the information capacity they can hide is relatively low, e.g., tens of bits. In order to hide high capacity watermarks like logo images and achieve better robustness, a novel deep invisible watermarking system is proposed. As shown in \Fref{fig:system_pipeline}, it consists of two main parts: one embedding sub-network $\mathbf{H}$ to learn how to hide invisible watermarks into the image, and another extractor sub-network $\mathbf{R}$ to learn how to extract the invisible watermark out. To avoid $\mathbf{R}$ generating watermark for all the images no matter whether they have invisible watermarks or not, we also constrain $\mathbf{R}$ not to extract any watermark out if its input is a clean image. Moreover, to make $\mathbf{R}$ generalize better for outputs of any surrogate model, an extra adversarial training stage is incorporated. In this stage, we pick one example surrogate model and add its outputs into $\mathbf{R}'s$ training. 
	
    For forensics, we leverage both the classic normalized correlation metric and a simple classifier to judge whether the extracted watermark is valid or not. Through extensive experiments, the proposed framework shows its strong ability in resisting the attack from surrogate models trained with different network structures like Resnet and UNet and different loss functions like $L1$, $L2$, perceptual loss, and adversarial loss. By jointly training the target model and watermark embedding, we find the functionality of $\mathbf{H}$ can be even absorbed into $\mathbf{M}$, making $\mathbf{M}$ a watermarked model itself.

	The contributions of this paper can be summarized in the following five aspects:

	\begin{itemize}
	    \item We introduce the IP protection problem for image processing networks. We hope it will help draw more attention to this seriously under-explored field and inspire more great works.
	    \item Motivated by the loss minimization property of deep networks, we innovatively propose to leverage the spatial invisible watermarking mechanism for deep model watermarking.
	    \item A novel deep invisible watermarking algorithm along with the two-stage training strategy is designed to improve the robustness and capacity of traditional spatial invisible watermarking methods.
	    \item We extend the proposed framework to multiple-watermark and self-watermark case, which further shows its strong generalization ability.
	    \item Both the classic normalized correlation metric and a watermark classifier demonstrate the strong robustness to surrogate model attack with different network structures and loss functions.
	  
	\end{itemize}
	
	Compared with the preliminary conference version \cite{zhang2020model}, we have made significant improvements and extension in this manuscript. The main difference are from four aspects: 1) We rewrite the theoretical analysis \Sref{sec:theo_ana} to make it more clear and solid; 2) We design a new classifier based watermark verification mechanism in \Sref{mt:wm_ver}; 3) We extend our framework to multiple watermarks and self-watermarked models in \Sref{mt:ext_mul} and \Sref{ext_swm}; 4) More corresponding experimental results and analysis are given in \Sref{sec:exp} and \Sref{sec:ext}.

	\section{Related Work}
	
	\subsection{Media Copyright Protection} 
	Compared to model IP protection, media copyright protection \cite{zhao1995embedding,nikolaidis1996copyright,lee1999adaptive,zafeiriou2005blind,lou2007copyright,fang2020deep} is a classic research field that has been studied for several decades. The most popular media copyright protection method is watermarking. For image watermarking, many different algorithms have been proposed in the past, which can be broadly categorized into two types: visible watermarks like logos, and invisible watermarks. Compared to visible watermarks, invisible watermarks are more secure and robust. They are often embedded in the original spatial domain \cite{kutter1999watermarking,voloshynovskiy2000content,voloshynovskiy2001multibit,deguillaume2002method}, or other image transform domains such as discrete cosine transform (DCT) domain \cite{hsu1999hidden,hernandez2000dct}, discrete wavelet transform (DWT) domain \cite{barni2001improved}, and discrete Fourier transform (DFT) domain \cite{ruanaidh1996phase}. However, all these traditional watermarking algorithms are often only able to hide several or tens of bits, let alone real logo images. More importantly, we find only spatial domain watermarking work to some extent for this task and all other transform domain watermarking algorithms totally fail.

	In recent years, some DNN-based watermarking schemes have been proposed. For example, Zhu \etal \cite{zhu2018hidden} propose an auto-encoder-based network architecture to realize the embedding and extracting process of watermarks. Based on it, Tancik \etal \cite{tancik2019stegastamp} further realize a camera shooting resilient watermarking scheme by adding a simulated camera shooting distortion to the noise layer. In \cite{deeba2020digital}, a full-size image is proposed to be used as watermark under the hiding image in image framework. Compared to these media watermarking algorithms, model watermarking is much more challenging because of the exponential search space of deep models. But we innovatively find it possible to leverage spatial invisible watermarking techniques for model protection.

	\subsection{Deep Model IP protection}
	Though IP protection for deep neural networks is still seriously under-studied, there are some recent works \cite{uchida2017embedding,adi2018turning,zhang2018protecting,nagai2018digital,chen2019blackmarks,darvish2019deepsigns,wu2020watermarking} that start paying attention to it. For example, based on the over-parameterized property of deep neural networks, Uchida \etal \cite{uchida2017embedding} propose a special weight regularizer to the objective function so that the distribution of weights can be resilient to attacks such as fine-tuning and pruning. One big limitation of this method is not task-agnostic and need to know the original network structure and parameters for retraining. Adi \etal \cite{adi2018turning} use a particular set of inputs as the indicators and let the model deliberately output specific incorrect labels, which is also known as ``backdoor attack". Similar scheme is adopted in \cite{zhang2018protecting} and \cite{guo2018watermarking} for DNN watermarking on embedded devices. Recently, Fan \etal \cite{fan2019rethinking} find all the above backdoor based methods are fragile to ambiguity attack. To overcome this limitation, a special digital passport layer is embedded into the target model and trained in an alternative manner. But this special passport layer often needs to make some network structure changes and causes performance degradation. So in the recent work \cite{zhang2020passport}, Zhang \etal further introduce a new passport-aware normalization layer for model IP protection.
	
	Despite the effectiveness of the above methods,  they all focus on the classification tasks, which is different from the purpose of this paper, i.e., protecting higher commercial valued image processing models. And for the challenging surrogate model attack, they will totally fail. In this paper, we consider the surrogate model attack  and study the IP protection for image processing networks. And different from the above methods, we innovatively propose to leverage spatial invisible watermarking algorithms for deep model IP protection.

	\subsection{Image-to-image Translation}
	In the deep learning era, most image processing tasks such as style transfer \cite{chen2017stylebank,chen2017coherent,chen2020explicit}, deraining\cite{chen2018gated,li2018recurrent,chen2020controllable}, and X-ray image debone \cite{yang2017cascade}, can be modeled as an image-to-image translation problem where both the input and output are images. Thanks to the emergence of the generative adversarial network (GAN) \cite{goodfellow2014generative}, this field has achieved significant progress recently. Isola \etal propose a general image-to-image translation framework Pix2Pix by combining adversarial training in \cite{pix2pix2017}, which is further improved by many following works \cite{choi2018stargan,wang2018high,park2019semantic} in terms of quality, diversity and functionality. However, all these methods  need a lot of pairwise training data, which is often difficult to be collected. By introducing the cycle consistency, Zhu \etal propose a general unpaired image-to-image translation framework CycleGAN \cite{zhu2017unpaired}. A similar idea is also adopted in concurrent works DualGAN \cite{yi2017dualgan} and DiscoGAN \cite{kim2017learning}. In this paper, we mainly focus on the deep models of paired image-to-image translation, because the paired training data is much more expensive to be obtained than unpaired datasets. More importantly, there is no prior work that has ever considered the watermarking issue for such models.

	\begin{figure}
	    \centering
	    \includegraphics[width=0.95\linewidth]{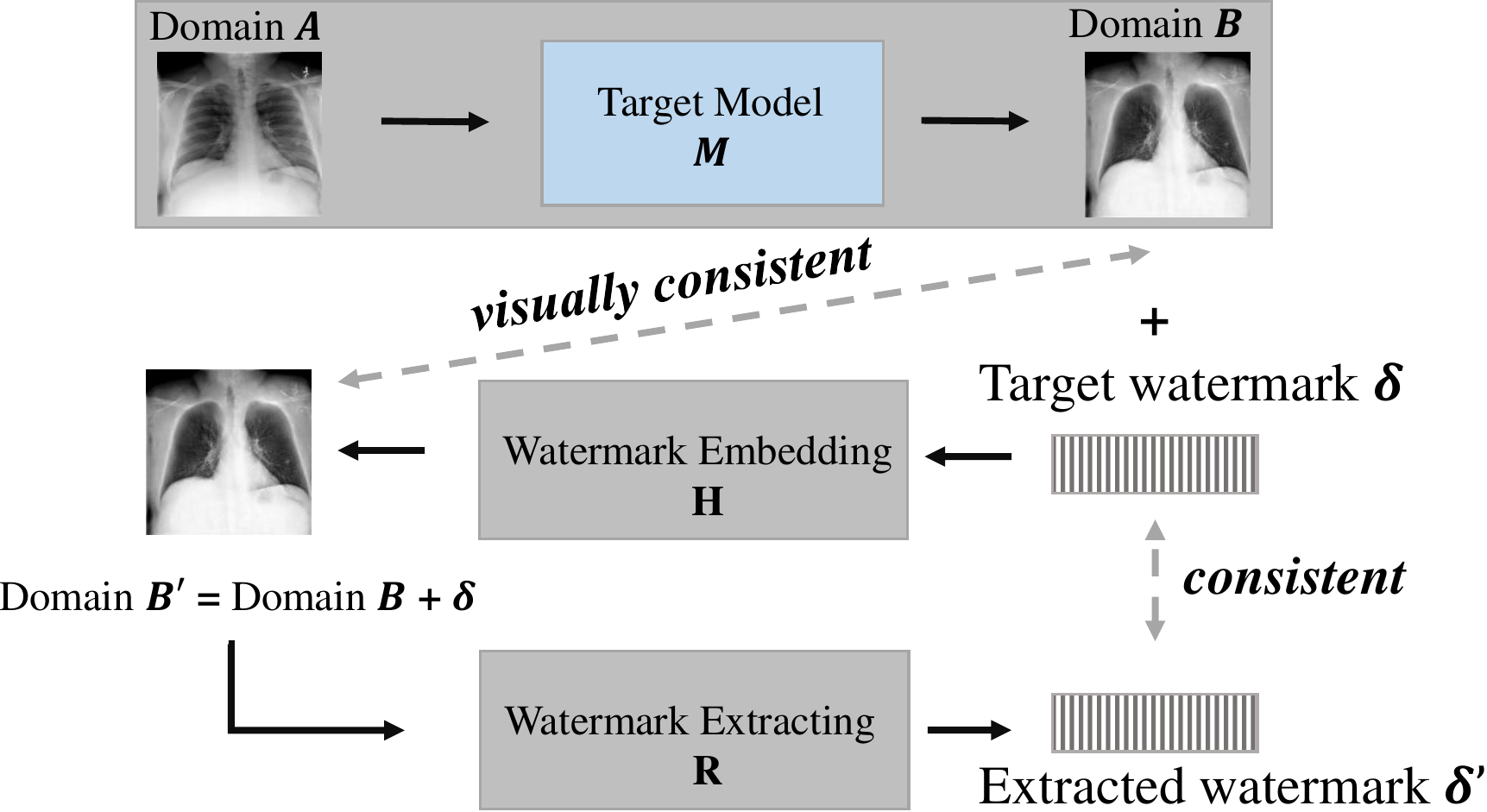}
	    \caption{The proposed deep watermarking framework by leveraging spatial invisible watermarking algorithms.}
	    \label{fig:framework}
	\end{figure}
	
	\section{Method}
	
	In this section, we will first define the target problem formally in \Sref{sec:prob_def}, then provide the theoretical analysis why the spatial invisible watermarking can be utilized for model watermarking in \Sref{sec:theo_ana}. Based on the analysis, traditional spatial invisible watermarking and a new deep invisible watermarking framework will be elaborated in \Sref{sec:trad_water} and \Sref{sec:deep_water} respectively. Next, two different watermark verification methods will be considered in \Sref{mt:wm_ver}. Finally, we will show how to extend the proposed framework to enable multiple watermarks within one single sub-network in \Sref{mt:ext_mul} and make the target model self-watermarked in \Sref{ext_swm}.

	\subsection{Problem Definition} \label{sec:prob_def}
	
    Given a target model $\mathbf{M}$ to protect, this paper mainly considers the surrogate model attack, where the attacker does not know the detailed network structure or weights of $\mathbf{M}$ but is able to access its output. This is a very common and sensible setting for real systems because most existing commercial deep models are deployed as cloud API service or encrypted executable program. In these scenarios, though the attacker cannot use common white-box attack methods like fine-tuning and pruning, a surrogate model can be trained to imitate the target model's behavior in a teacher-student way. Specifically, the attacker collects a lot of input images $\{a_1, a_2, ..., a_m\}$ (domain $\mathbf{A}$) and feeds them into $\mathbf{M}$ to get corresponding output images $\{b_1, b_2, ..., b_m\}$ (domain $\mathbf{B}$), then a surrogate model $\mathbf{SM}$ is trained in a supervised way by regarding $\mathbf{M}$'s outputs as ground truth.
	

	For effective forensics, the goal of resisting surrogate model attack is to design an effective way that is able to identify $\mathbf{SM}$ once it is trained with the data generated by $\mathbf{M}$. And in real scenarios, it is highly possible that we cannot access the detailed information of model $\mathbf{SM}$ (e.g., network structure and weights) like the target model neither. This means that, even when we suspect one model is a pirated model, the only indicator we can utilize is its output unless we start the costly legal proceeding. Therefore, we need to figure out one way to extract some kinds of forensics hints from the output of $\mathbf{SM}$. In this paper, we propose a new deep model watermarking mechanism so that the pre-defined watermark pattern can be extracted from $\mathbf{SM}$'s output.

	\subsection{Theoretical Pre-analysis.}\label{sec:theo_ana}
	In traditional watermarking algorithms, given an image $I$ and a target watermark ${\delta}$ to be embed, they will first use a watermark embedder $\mathbf{H}$ to generate an image $I'$ which contains ${\delta}$. Symmetrically, the target watermark ${\delta}$ can be further extracted out by the corresponding watermark extractor $\mathbf{R}$. As shown in \Fref{fig:simple_vis_wm}, if each image $b_i \in \mathbf{B}$ is embedded with a unified watermark ${\delta}$, forming another domain $\mathbf{B}'$, then the objective of the surrogate model $\mathbf{SM}$ is to minimize the distance $\mathcal{L}$ between $\mathbf{SM}(a_i)$ and $b_i'$:
	\begin{equation}
	\mathcal{L}(\mathbf{SM}(a_i), b_i')\rightarrow 0, \quad b_i' = b_i + \sigma.
	\end{equation}
	Below we will first show $\mathbf{SM}$ can learn the watermark $\delta$ into its output, then show the inherent loss minimization property of deep network will theoretically guarantee $\delta$ to be learned by a good surrogate model $\mathbf{SM}$.
	
	In detail, for each $a_i$ we have $b_i = M(a_i)$, thus there must exist a model $\mathbf{SM}$ that can learn good transformation between domain $\mathbf{A}$ and domain $\mathbf{B'}$ because of the existence of below equivalence. 

	\begin{equation}
	\begin{aligned}
	    \mathcal{L}(\mathbf{M}(a_i), b_i) & \rightarrow 0 \Leftrightarrow \mathcal{L}(\mathbf{SM}(a_i), (b_i+\mathbf{\delta})) \rightarrow 0 \\
	   \mbox{when} \quad \mathbf{SM} &= \mathbf{M} + \mathbf{\delta}.
	\end{aligned}
	\end{equation}
	In other words, one simplest solution of $\mathbf{SM}$ is to directly add ${\delta}$ to the output of $\mathbf{M}$ with a skip connection.
	
	On the other hand, because the objective of $\mathbf{SM}$ is calculated based on $b_i'$, the loss minimization property of deep network theoretically guarantees a good surrogate model $\mathbf{SM}$ should learn the unified watermark $\delta$ into its output. Otherwise, its objective loss function $\mathcal{L}$ cannot achieve a lower value than the above simple solution.

	Based on this observation, we propose a general deep watermarking framework for image processing models shown in \Fref{fig:framework}. Given a target model $\mathbf{M}$ to protect, we add a barrier $\mathbf{H}$ after it and embed a unified watermark ${\delta}$ into its output before showing them to the end-users. In this way, the final output obtained by the attacker is actually the watermarked version, so the surrogate model $\mathbf{SM}$ has to be trained with the image pair $(a_i, b'_i)$ from domain $\mathbf{A},\mathbf{B'}$ instead of the original pair $(a_i, b_i)$ from domain $\mathbf{A},\mathbf{B}$. After $\mathbf{SM}$'s training, we can leverage the corresponding watermark extractor $\mathbf{R}$ to extract the watermark out from the output of $\mathbf{SM}$.

	To ensure the watermarked output image $b'_i$ is visually consistent with the original one $b_i$, only spatial invisible watermarking algorithms are considered in this paper. Below we will introduce both traditional spatial invisible watermarking algorithm and a novel deep invisible watermarking algorithm.

	\subsection{Traditional Spatial Invisible Watermarking.}\label{sec:trad_water} Additive-based embedding is the most common method used in the traditional spatial invisible watermarking scheme. Specifically, the watermark information is first spread into a sequence or block which satisfies a certain distribution, then embedded into the corresponding coefficients of the host image. This embedding procedure can be formulated by
	\begin{equation}\label{TraSpatialMethod}
	    I^{\prime}=\left\{\begin{array}{ll}{I+\alpha C_{0}} & {\text { if } w_i=0} \\ {I+\alpha C_{1}} & {\text { otherwise }}\end{array}\right.
	\end{equation}
	where $I$ and $I^{\prime}$ indicate the original image and embedded image respectively. $\alpha$ indicates the embedding intensity and $C_i$ denote the spread image block that represents bit ``$w_i$"($w_i\in[0,1]$). In the extraction side, the watermark is determined by detecting the distribution of the corresponding coefficients. The robustness of such an algorithm is guaranteed by the spread spectrum operation. The redundancy brought by the spread spectrum makes a strong error correction ability of the watermark so that the distribution of the block will not change a lot even after image processing. 

	However, such algorithms often have very limited embedding capacity because many extra redundant bits are needed to ensure robustness. In fact, in many application scenarios, the IP owners may want to embed some special images (e.g., logos) explicitly for convenient visual forensics. This is nearly infeasible for these algorithms. More importantly, the following experiments show that these traditional algorithms can only resist the attack from some special types of surrogate models. To enable more high-capacity watermarks and more robust resistance ability, we will propose a new deep invisible watermarking algorithm below.

	\begin{figure*}[!ht]
	\centering
	\includegraphics[width=0.95\linewidth]{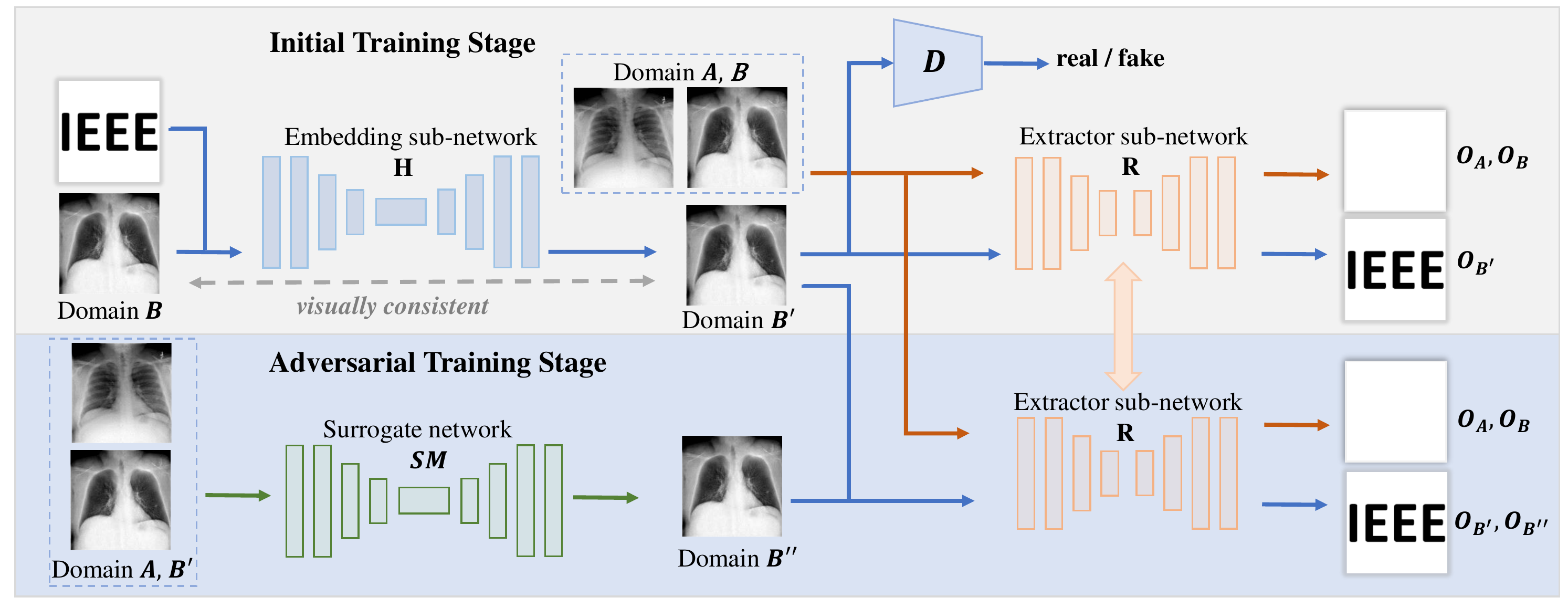}
	\vspace{-1em}
	\caption{The overall pipeline of the proposed deep invisible watermarking algorithm and two-stage training strategy. In the first training stage, a basic watermark embedding sub-network $\mathbf{H}$ and extractor sub-network $\mathbf{R}$ are trained. Then another surrogate network $\mathbf{SM}$ is leveraged as the adversarial competitor to further enhance the extracting ability of $\mathbf{R}$.}
	\label{fig:system_pipeline}
	\end{figure*}

	\subsection{Deep Invisible Watermarking.}\label{sec:deep_water} \subsubsection{Overview} As shown in \Fref{fig:system_pipeline}, to embed an image watermark into host images of the domain $\mathbf{B}$ and extract it out afterward, one embedding sub-network $\mathbf{H}$ and one extractor sub-network $\mathbf{R}$ are adopted respectively. Without sacrificing the original image quality of domain $\mathbf{B}$, we require watermarked images in domain $\mathbf{B'}$ to be visually consistent to the original images in the domain $\mathbf{B}$. As adversarial networks have demonstrated their power in reducing the domain gap in many image translation tasks, we append one discriminator network $\mathbf{D}$ after $\mathbf{H}$ to further improve the image quality of domain $\mathbf{B'}$.
	
	During training, we find if the extractor network $\mathbf{R}$ is only trained with the images of domain $\mathbf{B'}$, it is very easy to overfit and output the target watermark no matter whether the input images contain watermarks or not. To circumvent this problem, we also feed the images of domain $\mathbf{A}$ and domain $\mathbf{B}$ that do not contain watermarks into $\mathbf{R}$ and force it to output a constant blank image. In this way, $\mathbf{R}$ will have the real ability to extract watermarks out only when the input image has the watermark in it.  

    Based on the pre-analysis, when the attacker uses a surrogate model $\mathbf{SM}$ to imitate the target model $\mathbf{M}$ based on the input domain $\mathbf{A}$ and watermarked domain $\mathbf{B'}$, $\mathbf{SM}$ will learn the embedded watermark ${\delta}$ into its output. Despite higher embedding capacity, we find that the extractor sub-network $\mathbf{R}$ cannot extract the watermarks out from the output of the surrogate model $\mathbf{SM}$ neither like traditional watermarking algorithms if only trained with this initial training stage. This is because $\mathbf{R}$ has only observed clean watermarked images but not the watermarked images from surrogate models which may contain some unpleasant noises. To further enhance the extracting ability of $\mathbf{R}$, we choose one simple surrogate network to imitate the attackers' behavior and fine-tune $\mathbf{R}$ on the mixed dataset of domain $\mathbf{A}, \mathbf{B}, \mathbf{B'}, \mathbf{B''}$ in an extra adversarial training stage. The following experiments will show this can significantly boost the extracting ability of $\mathbf{R}$. More importantly, this ability generalizes well to other types of surrogate models.

	\subsubsection{Network Structures}
	By default, we adopt UNet \cite{ronneberger2015u} as the  network structure of $\mathbf{H}$ and $\mathbf{SM}$, which has been widely used by many translation based tasks like Pix2Pix \cite{pix2pix2017} and CycleGAN\cite{zhu2017unpaired}. It performs especially well for tasks where the output image shares some common properties of input image by using multi-scale skip connections. But for the extractor sub-network $\mathbf{R}$ whose output is different from the input, we find CEILNet \cite{fan2017generic} works much better. It also follows an auto-encoder like network structure. In details, the encoder consists of three convolutional layers, and the decoder consists of one deconvolutional layer and two convolutional layers symmetrically. In order to enhance the learning capacity, nine residual blocks are inserted between the encoder and decoder. For the discriminator $\mathbf{D}$, we adopt the widely-used PatchGAN \cite{pix2pix2017}. Note that except for the extractor sub-network $\mathbf{R}$, we find other types of translation networks also work well in our framework, which demonstrates the strong generalization ability of our framework.

	\subsubsection{Loss Functions}
	The objective loss function of our method consists of two parts: the embedding loss $\mathcal{L}_{emd}$ and the extracting loss $\mathcal{L}_{ext}$, i.e.,
	\begin{equation}
	    \mathcal{L} = \mathcal{L}_{emd} + \lambda*\mathcal{L}_{ext},
	\end{equation}
	where $\lambda$ is the hyper parameter to balance these two loss terms. Below we will introduce the detailed formulation of $\mathcal{L}_{emd}$ and $\mathcal{L}_{ext}$ respectively.

    \vspace{1em}
	\noindent\textbf{Embedding Loss.}
	To embed the watermark image into a cover image while guaranteeing the original visual quality, three different types of visual consistency loss are considered: the basic $L$2 loss $\ell_{bs}$, perceptual loss $\ell_{vgg}$ and adversarial loss $\ell_{adv}$, i.e.,
	\begin{equation}
	\begin{aligned}
	    \mathcal{L}_{emd} &= \lambda_1 * \ell_{bs} + \lambda_2 * \ell_{vgg} + \lambda_3 * \ell_{adv}.
	\end{aligned}
	\end{equation}
	Here the basic $L$2 loss $\ell_{bs}$ is simply the pixel value difference between the input host image $b_i$  and  the watermarked output image $b_i'$, $N_c$ is the total pixel number, i.e.,
	\begin{equation}
	\begin{aligned}
	    \ell_{bs} &= \sum_{b_i' \in \mathbf{B}', b_i \in \mathbf{B}} \frac{1}{N_c} \lVert b_i' - b_i\rVert^2. \\
	\end{aligned}
	\end{equation}
	And the perceptual loss $\ell_{vgg}$ \cite{johnson2016perceptual} is defined as the difference between the VGG feature of $b_i$ and $b_i'$:
	\begin{equation}
	\begin{aligned}
	    \ell_{vgg} &= \sum_{b_i' \in \mathbf{B}', b_i \in \mathbf{B}} \frac{1}{N_f} \lVert VGG_k(b_i') - VGG_k(b_i)\rVert^2,
	\end{aligned}
	\end{equation}
	where $VGG_k(\cdot)$ denotes the features extracted at layer $k$ (``conv2\_2" by default), and $N_f$ denotes the total feature neuron number. To further improve the visual quality and minimize the domain gap between $\mathbf{B}'$ and $\mathbf{B}$, the adversarial loss $\ell_{adv}$ will let the embedding sub-network $\mathbf{H}$ embed watermarks better so that the discriminator $\mathbf{D}$ cannot differentiate its output from real watermark-free images in $\mathbf{B}$.
	\begin{equation}
	\begin{aligned}
	    \ell_{adv} &= \underset{b_i\in \mathbf{B}}{\mathbb{E}} log(\mathbf{D}(b_i)) + \underset{b_i'\in \mathbf{B'}}{\mathbb{E}} log(1 - \mathbf{D}(b_i')). \\
	\end{aligned}
	\end{equation}
    
     \vspace{1em}
	\noindent\textbf{Extracting Loss.} The responsibility of the extractor sub-network $\mathbf{R}$ has two aspects: it should be able to extract the target watermark out for watermarked images from $\mathbf{B'}$ and instead output a constant blank image for watermark-free images from $\mathbf{A,B}$. So the first two terms of $\mathcal{L}_{ext}$ are the reconstruction loss $\ell_{wm}$ and the clean loss $\ell_{clean}$ for these two types of images respectively, i.e.,
	\begin{equation}
	\label{eq:wm}
	    \begin{aligned}
	    \ell_{wm} &= \sum_{b_i' \in \mathbf{B}'}\frac{1}{N_c}\lVert R(b_i') - \delta\rVert^2,  \\
	    \ell_{clean} = \sum_{a_i \in \mathbf{A}}\frac{1}{N_c}\lVert &R(a_i) - \delta_0\rVert^2 + \sum_{b_i \in \mathbf{B}}\frac{1}{N_c}\lVert R(b_i) - \delta_0\rVert^2, \\
	    \end{aligned}
	\end{equation}		
	where $\delta$ is the target  watermark image and $\delta_0$ is the constant blank watermark image. Besides the reconstruction loss, we also want the watermarks extracted from different watermarked images to be consistent, thus another consistent loss $\ell_{cst}$ is added:
	\begin{equation}
	\label{eq:cst}
	    \ell_{cst} = \sum_{x,y \in \mathbf{B'}} \lVert R(x) - R(y)\rVert^2.
	\end{equation}	
	The final extracting loss $\mathcal{L}_{ext}$ is defined as the weighted sum of these three terms, i.e.,
	\begin{equation}
	    \mathcal{L}_{ext} = \lambda_4*\ell_{wm} + \lambda_5*\ell_{clean} + \lambda_6*\ell_{cst}.
	\end{equation}

    \vspace{1em}
	\noindent\textbf{Adversarial Training Stage.} From experiments, we find that if $\mathbf{R}$ is only trained with the above initial training stage, it cannot generalize well for the noisy watermarked output of some surrogate models because only the clean watermarked images are observed. To enhance its extracting ability, an extra adversarial training stage is added, wherein we introduce the degradation of surrogate model attack. Specifically, one proxy surrogate model $\mathbf{SM}$ is trained with the simple $L$2 loss by default. Denote the outputs of this proxy model as $\mathbf{B}''$, we further fine-tune $\mathbf{R}$ on the mixed dataset $\mathbf{A, B, B', B''}$ in this stage. Since the embedding sub-network $\mathbf{H}$ is fixed in this stage, only the extracting loss $\mathcal{L}_{ext}$ is considered to refine the capability of $\mathbf{R}$. Specifically, the clean loss $\ell_{clean}$ is kept unchanged, while $\ell_{wm}, \ell{cst}$ are updated as below respectively:
	\begin{equation}
	\begin{aligned}
	 \ell_{wm} &= \sum_{b_i' \in \mathbf{B}'}\frac{1}{N_c}\lVert R(b_i') - \delta\rVert^2 + \sum_{b_i'' \in \mathbf{B}''}\frac{1}{N_c}\lVert R(b_i'') - \delta\rVert^2, \\
	  \ell_{cst} &= \sum_{x,y \in \mathbf{B'}\cup\mathbf{B''}} \lVert R(x) - R(y)\rVert^2.  
	 \end{aligned}
	\end{equation}
	
	\subsubsection{Watermark Verification}\label{mt:wm_ver}
	For effective forensics, if we want to know whether a surrogate model $\mathbf{SM}$ is trained by stealing the IP of one target model, we first feed $\mathbf{SM}$'s output into the extractor sub-network $\mathbf{R}$, then verify whether $\mathbf{R}$'s output matches the pre-defined watermark. In this paper, we consider two different verification methods: the classic normalized correlation (NC) metric and an extra classifier $\mathbf{C}$. For the former one, it is simply defined as: 
	\begin{equation}
	NC = \frac{<\mathbf{R}(b_i'),\delta>}{\lVert \mathbf{R}(b_i')\rVert*\lVert\delta\rVert},
	\end{equation}
	where $<\cdot, \cdot>$ and $\lVert\cdot\rVert$ denote the inner product and L2 norm respectively. Compared to the NC metric, the latter classifier is more straightforward and robust to noisy parts of $\mathbf{R}(b_i')$ potentially. Because it is just a binary classification problem (yes/no), we find a simple network consisting of three convolutional layers is enough. In our method, $\mathbf{C}$ is joint trained with $\mathbf{R}$ during the adversarial training stage, and label ``1" and ``0" represent watermarked and watermark-free images respectively. By default, the simple cross entropy loss is used:

	\begin{small}
	\begin{equation}
	    \mathcal{L}_{cls} = -[\sum_{x\in \mathbf{A}\cup\mathbf{B}} log(1-\mathbf{C}(x)) + \sum_{x\in\mathbf{B'}}log(\mathbf{C}(x))].\\
	\end{equation}
	\end{small}

\subsubsection{Extension to Multiple Watermarks}\label{mt:ext_mul}
In our default setting, for one specific watermark, an embedding sub-network and a corresponding extractor sub-network will be trained. This makes sense for the scenarios where only one specific deep model needs to be protected. However, it is unfriendly to some other scenarios where one deep model has multiple release versions and different versions are represented by different watermarks, because training multiple embedding and extractor sub-networks is both storage- and computation-consuming. Thanks to the flexibility of the proposed framework, enabling multiple different watermarks within one single embedding and extractor sub-network is easy to be supported. The overall training strategy is similar, and the biggest difference is that different watermarks will be randomly selected for $\mathbf{H}$ to embed and $\mathbf{R}$ to exact correspondingly. For the verification classifier $\mathbf{C}$, it will be changed from the binary classification to multi-class classification.

\subsubsection{Extension to Self-watermarked Models}\label{ext_swm}
As mentioned before, in order to protect one specific target model, the embedding sub-network acts as an extra barrier and will be appended after the target model as a whole. The main benefit of this pipeline is task-agnostic, which means that it is a very general solution and can work independently without the need of knowing the information of the target model. However, requiring an extra embedding sub-network may sometimew be a limitation. So we propose an extended solution to absorb the watermarking functionality into the target model itself if we can control its training process. In this sense, the target model is self-watermarked without the extra embedding sub-network. In other words, the target model itself will automatically embed a watermark into its output.

\begin{figure}[t]
	\centering
	\includegraphics[width=0.98\linewidth]{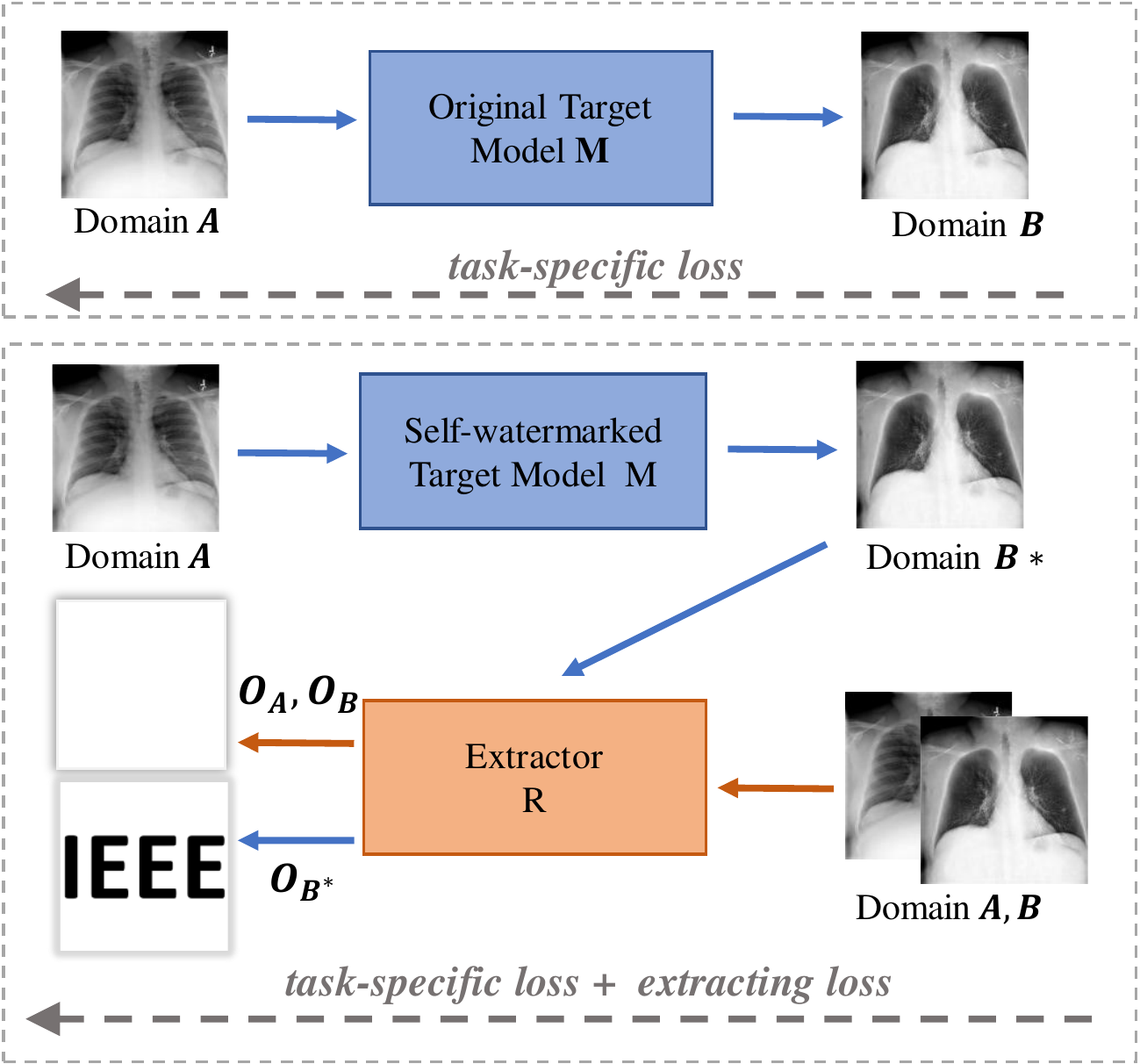}
	\caption{Illustration diagram to show the training difference between the original target model and self-watermarked target model.}
	\label{fig:task_spe_dd}
\end{figure}

The training difference between the original target model and the self-watermarked target model is shown in \Fref{fig:task_spe_dd}. Assuming the original target model $\mathbf{M}$ is trained with the task-special loss $\mathcal{L}_{task}$, we add an extractor sub-network $\mathbf{R}$ after $\mathbf{M}$ and jointly train them by combining the extracting loss $\mathcal{L}_{ext}$ mentioned above with $\mathcal{L}_{task}$. To ensure the robustness and capability of $\mathbf{R}$, $\mathbf{R}$ still needs to be fine-tuned with the extra adversarial training stage.

	\begin{figure*}[t]
		\centering
		\includegraphics[width=\linewidth]{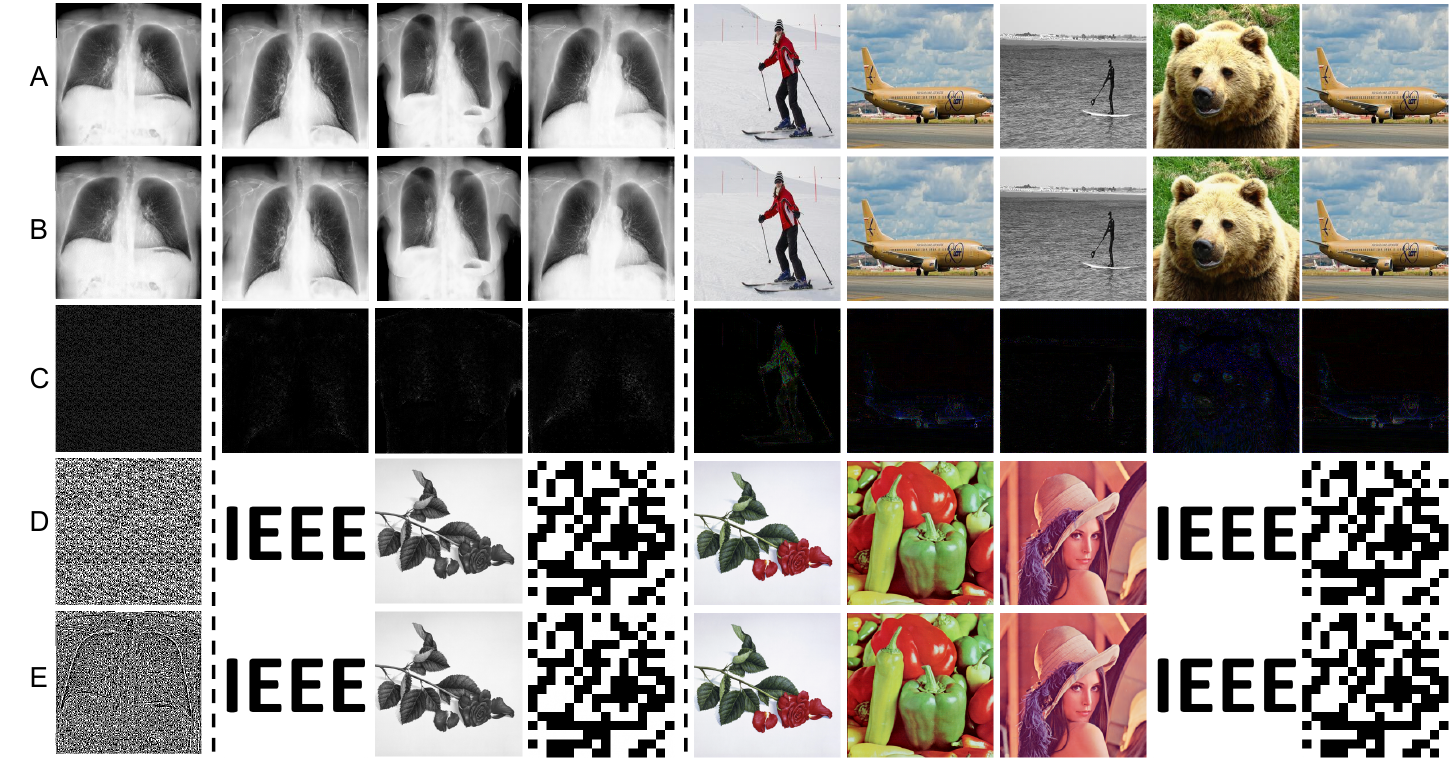}
		\caption{Some visual examples to show the capability of the proposed deep invisible watermarking algorithm: (A) watermark-free image $b_i$ from domain $\mathbf{B}$, (B) watermarked image $b_i'$ from domain $\mathbf{B'}$  (C) the residual between $b_i$  and $b_i'$ (enhanced 10$\times$), (D) ground-truth watermark, (E) extracted watermark  from $b_i'$. The first column is the results of traditional spatial bit-based invisible watermarking algorithms (64-bit embedded), and the middle and right parts are the results of our methods for the debone and deraining tasks respectively.}
		\label{fig:invisible_wm_qual}
	\end{figure*}
	
	\section{Experiments}\label{sec:exp}
	To demonstrate the effectiveness of the proposed system, we use two example image processing tasks in this paper: image deraining and Chest X-ray image debone. The goal of these two tasks is to remove the rain streak and rib components from the input images respectively. We will first introduce the implementation details and evaluation metric in \Sref{sec:exp_detail}, then evaluate the proposed deep invisible watermarking algorithm quantitatively and qualitatively in \Sref{sec:exp_invis_wm}. Next, we will demonstrate the robustness of the proposed deep watermarking framework to surrogate models with different network structures and loss functions in \Sref{sec:exp_rob}. Furthermore, we compare our method with traditional and DNN-based watermarking algorithm in \Sref{exp:cmp2} and  show the robustness of our method to watermark overwriting in \Sref{exp:ow}. Finally, some ablation analysis are provided in \Sref{exp:ab_study}.	
	
	\subsection{Implementation Details} \label{sec:exp_detail}

    \noindent\textbf{Dataset setup.} For image deraining, we use 6100 images from the PASCAL VOC dataset \cite{everingham2010pascal} as target domain $\mathbf{B}$, and use the synthesis algorithm in \cite{zhang2018density} to generate rainy images as domain $\mathbf{A}$. These images are split into two parts: 6000  are for both the initial and adversarial training stage and 100 for testing. Since the images used by the attacker may be different from that used by the IP owner to train $\mathbf{H,R}$, we simulate it by choosing 6000 images from the COCO dataset \cite{lin2014microsoft} for the surrogate model training. Similarly, for X-ray image debone, we select 6100 high-quality chest X-ray images from the open dataset chestx-ray8 \cite{wang2017chestx} and use the rib suppression algorithm proposed by \cite{yang2017cascade} to generate the training pair. They are divided into three parts: 3000 for both the initial and adversarial training, 3000 for the surrogate model training, and 100 for testing.  All the images are resized to 256 $\times$ 256 by default.

	\vspace{1em}
    \noindent \textbf{Training details.} By default, we train sub-network $\mathbf{H,R}$ and discriminator $\mathbf{D}$ for 200 epochs with a batchsize of 8. Adam optimizer is adopted with the initial learning rate of 0.0002. We decay the learning rate by 0.2 if the loss does not decrease within 5 epochs. For surrogate model training,  batchsize of 16 and longer epochs of 300 are used to ensure better performance. And the initial learning rate is set to be 0.0001, which stays unchanged in the first 150 epochs and is linearly decayed to zero in the remaining 150 epochs. In the adversarial training stage, sub-network  $\mathbf{R}$ is trained with the initial learning rate of 0.0001.  For classifier $\mathbf{C}$, it is also trained in a similar way.  By default, all hyperparameters $\lambda_i$ equal to 1 except $\lambda_3=0.01$.
	
	\vspace{1em}
    \noindent\textbf{Evaluation Metric.} To evaluate the visual quality, PSNR and SSIM are used by default. When using the NC metric for watermark verification, the watermark is regarded as successfully extracted if its NC value is bigger than $0.95$. Based on it, the success rate  ($SR_{NC}$) is further defined as the ratio of watermarked images whose hidden watermark is successfully extracted in an image set. When using the classifier for verification, we use the threshold 0.5 for the binary case and adopt the label with max-possibility for the multi-class case. Similarly, the success rates based on the classifier are denoted as $SR_C$.

	\begin{figure}[h]
	\centering
	\includegraphics[width=0.8\linewidth]{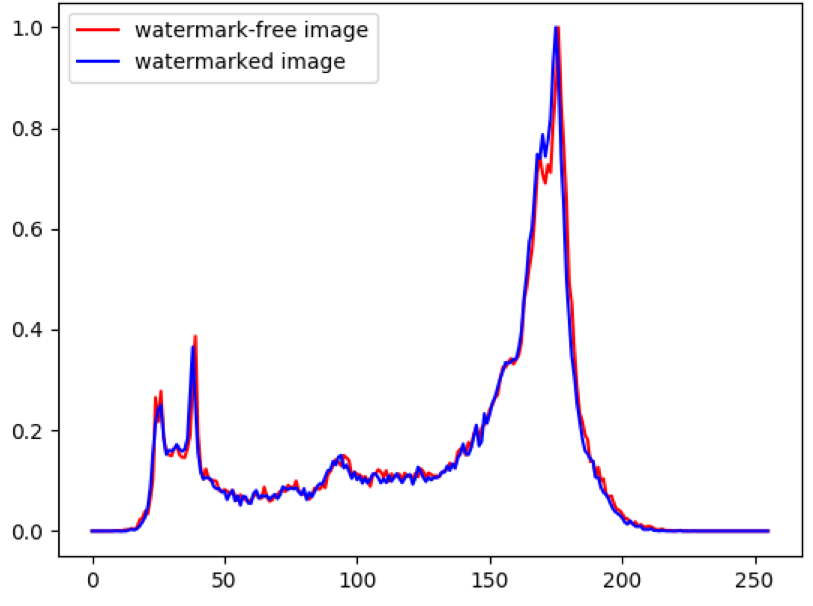}
	\vspace{-1em}
	\caption{The gray histogram comparison between watermark-free image and watermarked image for the Derain-Flower case.}
	\label{fig:hist}
	\end{figure}

\subsection{Deep Image Based Invisible Watermarking} \label{sec:exp_invis_wm}
	In this experiment, we give both qualitative and quantitative results about the proposed deep image based invisible watermarking algorithm. To demonstrate its generalization ability to various types of watermark images, both simple grayscale logo images and complex color images are considered. For the grayscale case, we even consider a QR code image. Several example images used for the debone and deraining task are shown in \Fref{fig:invisible_wm_qual}.  It can be easily seen that the proposed deep watermarking algorithm can not only embed the watermark images into the host images in an imperceptible way but also successfully extract the embedded watermarks out. Especially for the challenging ``Peppers" and ``Lena" watermarks that have rich textures, the embedding sub-network and extractor sub-network still collaborate very well and guarantee the visual quality of the watermarked image and extracted watermark simultaneously. 
	
	To investigate whether the extracting sub-network embeds watermark in a naive additive way, we further show the residuals (even enhanced 10 $\times$) between watermarked image $b_i'$ and  watermark-free image $b_i$ in \Fref{fig:invisible_wm_qual}. Obviously, we cannot observe any watermark hint from the residuals, which means the watermark is indeed embedded in an advanced and smart way. Taking one more step forward, we use the ``Flower" watermark image as an example and show the color histogram distribution difference between the watermarked image and original watermark-free image in \Fref{fig:hist}. It shows that, even though the colorful ``Flower" watermark image is embedded, the gray histogram distribution is almost identical to that of the original watermark-free image.
	
	Quantitative results with respect to the visual quality and extracting ability are provided in \Tref{tab:invisible_wm_quan}. We can find that the average PSNR and SSIM between watermarked and watermark-free images are very high, which double confirms the original image quality is well kept. On the other hand, though high extracting ability is contradictory to high visual quality to some extent, our extractor sub-network $\mathbf{R}$ is still able to achieve very high extracting ability with an average NC value over 0.99 and 100\% success rate both based on the NC metric and the classifier $\mathbf{C}$.

	\begin{table}[]
	\centering
	\caption{\label{tab:invisible_wm_quan} Quantitative results of the proposed deep  image based invisible watermarking algorithm. PSNR and SSIM are calculated between the watermarked and original watermark-free images. $x,y$ denote task name and watermark image name in the notation ``x-y".}
	\setlength{\tabcolsep}{2.8mm}{
	\begin{tabular}{c|c|c|c|c}
		\hline
		Task & PSNR & SSIM & NC &  $SR_{NC}$  $/$  $SR_C$\\
		\hline
		\hline
		Debone-IEEE & 47.29 & 0.99 & 0.9999 &100\%   $/$ 100\%          \\
		\hline
		Debone-Flower & 46.36 & 0.99 & 0.9999  &100\%   $/$ 100\%         \\
		\hline
		Debone-QR & 44.35 & 0.99 & 0.9999 &100\%   $/$ 100\%    \\
		\hline	
		\hline                  
		Derain-Flower & 41.21 & 0.99 & 0.9999  &100\%   $/$ 100\%     \\
		\hline
		Derain-Peppers & 40.91 & 0.98 & 0.9999  &100\%   $/$ 100\%     \\
		\hline
		Derain-Lena & 42.50 & 0.98 & 0.9999   &100\%   $/$ 100\%    \\
		\hline
		Derain-IEEE & 41.76 & 0.98 & 0.9999  &100\%   $/$ 100\%    \\
		\hline	
		Derain-QR & 40.24 & 0.98 & 0.9999  &100\%   $/$ 100\%    \\
		\hline
		\end{tabular}}		                  	         
	\end{table}
	
\begin{table*}[]
	\scriptsize
	\centering
	\caption{ The success rate ($SR_{NC} / SR_{C}$) of our method resisting the attack from surrogate models. Column 2 $\sim$ 5 are trained with $L$2 loss but different network structures and Column 6 $\sim$ 10 are trained with UNet network structure but different loss combinations. We take Debone-IEEE task and Derain-Flower task as example, and $\dagger$ denotes the results without adversarial training.}
	\label{tab:robustness}
	\setlength{\tabcolsep}{1.8mm}{ 	    
	\begin{tabular}{c|c|c|c|c|c|c|c|c|c}
	\hline		
         Settings      & CNet & Res9 & Res16 & UNet & $L$1 & $L$1 + $L_{adv}$ & $L$2 & $L$2 + $L_{adv}$ & $L_{perc}$+$L_{adv}$ \\ \hline\hline
         Debone & 93\%  $/$ 100\% & 100\%  $/$ 100\% & 100\%  $/$ 100\% & 100\%  $/$ 100\% & 100\%  $/$ 100\% & 100\%  $/$ 100\%  & 100\%  $/$ 100\%  & 100\% $/$ 100\%  & 71\% $/$ 94\%  \\
         \hline
         Derain & 100\%  $/$ 100\% & 100\%  $/$ 100\% & 100\%  $/$ 100\% & 100\%  $/$ 100\% &   100\%  $/$ 100\% & 100\%  $/$ 100\%  & 100\%  $/$ 100\%  & 99\% $/$ 100\% & 100\% $/$ 100\%  \\
         \hline
         Debone$\dagger$  &0\% $/$ 0\% &0\% $/$ 0\% &0\% $/$ 0\% & 0\% $/$ 0\% &0\% $/$ 0\%   &0\% $/$ 0\%   &27\% $/$ 88\%   &44\% $/$ 96\%   &0\% $/$ 0\%   \\
         \hline
         Derain$\dagger$  &0\% $/$ 0\% &0\% $/$ 0\% &0\% $/$ 0\% &0\% $/$ 0\%  &0\% $/$ 0\%  &0\% $/$ 0\%  &0\% $/$ 0\%   &0\% $/$ 0\%   &0\% $/$ 54\%  \\
         \hline
	\end{tabular}}
\end{table*}

	\begin{table*}[]
	    \scriptsize
	    \centering
	    \caption{ The comparison of embedding \& extracting ability (Column 2 $\sim$ 4) and robustness to surrogate model attack (Column 5 $\sim$ 8) among our method and two typical methods in debone task.  TSIW-64  denotes using traditional spatial invisible watermarking algorithms \cite{voloshynovskiy2001multibit} to hide 64-bit, while HiDDeN is a typical DNN-based watermarking method. We use HiDDeN to hide 64-bit as well and $\ddagger$ denotes training  HiDDeN without noise layer. The surrogate models are trained with $L$2 loss but different network structures. For TSIW-64 and HiDDeN, only $SR_{NC}$ is reported.}
	    \label{tab:cmp2}	
	    \setlength{\tabcolsep}{5.2mm}{    
    \begin{tabular}{c||c|c|c||c|c|c|c}
        \hline
         Method & PSNR & SSIM & NC & CNet & Res9 & Res16 & UNet \\ \hline\hline
         HiDDeN \cite{zhu2018hidden} &36.25 &0.94 & 0.7220 & 0\%  &0\%  & 0\%  &0\% \\
         \hline
         HiDDeN \cite{zhu2018hidden}$\ddagger$  & 37.23 & 0.97 &0.9972 & 0\%  &0\%  & 0\%  &0\% \\
         \hline
         TSIW-64 \cite{voloshynovskiy2001multibit} & 40.81 & 0.95 & 0.9999  & 0\%  &0\% & 0\%  &100\% \\
         \hline
         Our  &47.29  &0.99 & 0.9999   & 93\%  $/$ 100\% & 100\%  $/$ 100\% & 100\%  $/$ 100\% & 100\%  $/$ 100\%  \\
         \hline
    \end{tabular}}
	\end{table*}

	\subsection{Robustness to Surrogate Model Attack} \label{sec:exp_rob}
	Besides the visual quality consistency, another more important goal is to guarantee IP protection ability. Considering the attacker may use different network structures trained with different loss functions to imitate target model's behavior,  we simulate this case by using a lot of surrogate models to evaluate the robustness of the proposed deep  watermarking framework. In details, we consider four different types of network structures: vanilla convolutional networks only consisting of several convolutional layers (``CNet"), an auto-encoder like networks with 9 and 16 residual blocks (``Res9", ``Res16"), and the aforementioned UNet network (``UNet"). For the objective loss function, popular loss functions like $L$1, $L$2, perceptual loss $L_{perc}$, adversarial loss $L_{adv}$ and their combination are considered. But we discard the case that only utilizes perceptual loss $L_{perc}$ for surrogate model learning because it will generate very terrible image quality (PSNR:19.73; SSIM:0.85) and make the attack meaningless. As the surrogate model with ``UNet" and $L$2 loss function is utilized in the adversarial training stage, this configuration can be viewed as a white-box attack and all other configurations are black-box attacks. Without losing generality,  we use grayscale ``IEEE" and color ``Flower" image as default target watermark for debone and deraining respectively in this experiment.
		
	\begin{figure*}[ht]
	    \centering
	    \includegraphics[width=0.95\linewidth]{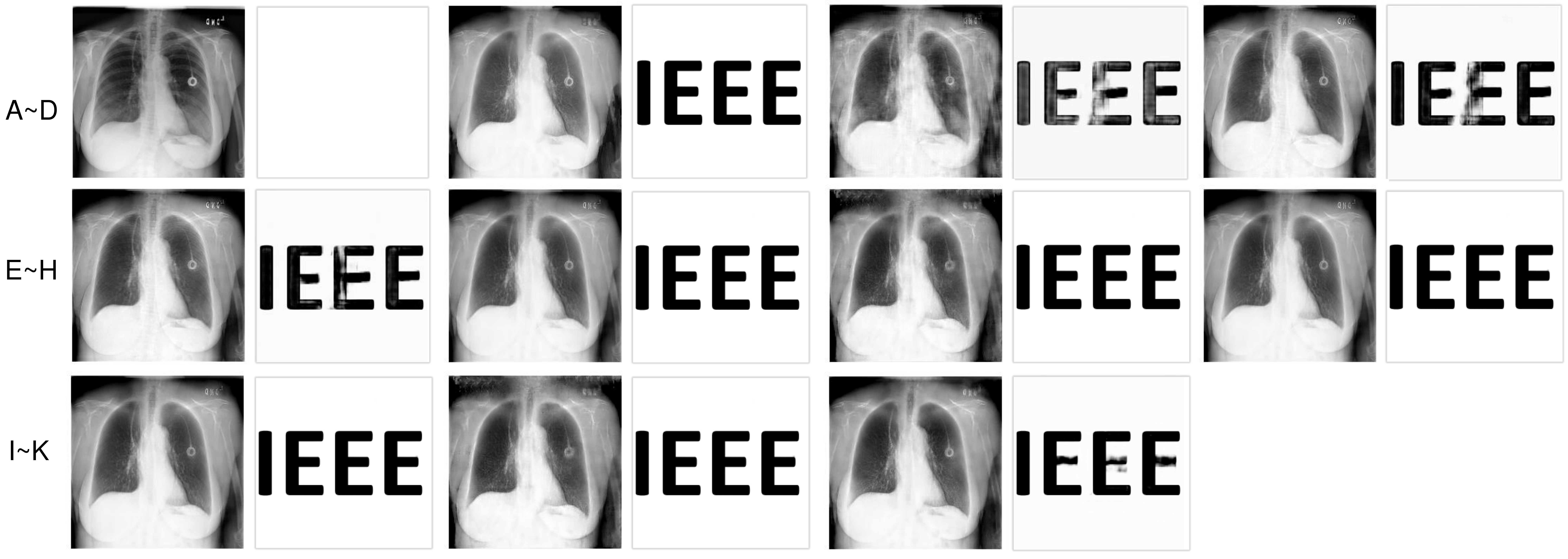}
	    \caption{Example output image and corresponding extracted watermark from different surrogate models. (A) input watermark-free image $a_i$; (B) watermarked image $b_i'$; (C) $\sim$ (F) are cases for different network structures with $L$2 loss: CNet, Res9, Res16 and UNet in turn; (G) $\sim$ (K) are cases for different loss combinations with UNet:  $L$2, $L$2+$L_{adv}$, $L$1, $L$1+$L_{adv}$ and $L_{perc}$+$L_{adv}$  in turn.}
	    \label{fig:vis_sr}
	\end{figure*}

	\begin{figure}[t]
	\centering
	\includegraphics[width=0.98\linewidth]{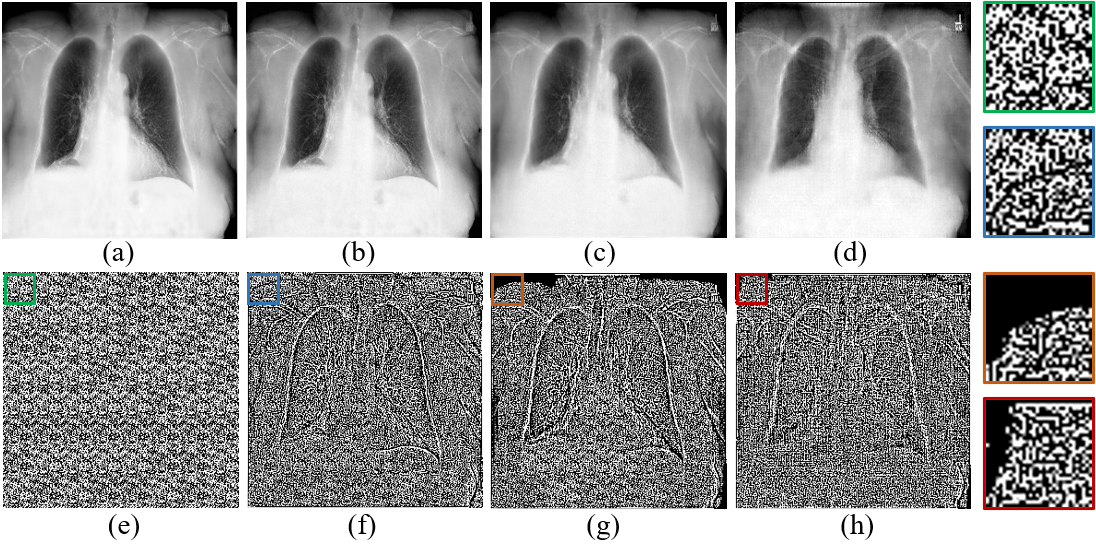}
	\caption{Visual results of traditional spatial invisible watermarking: (a) watermark-free image  $b_i$, (b) watermarked image $b_i'$, (c) and (d) are the outputs of surrogate model (UNet and Res9 with $L$2, respectively), (e) ground-truth watermark, (f) $\sim$ (h) are the corresponding extracted watermark from (b) $\sim$ (d).  It is defined as successful extracting when the extracted pattern matches mostly with ground-truth pattern like pattern in blue and orange box, otherwise as failure like pattern in red box. }
	\label{fig:t-sr}
	\vspace{-1em}
	\end{figure}

	Because of the limited computation resource, we do not consider all the combinations of network structures and loss functions. Instead, we choose to conduct the control experiments to demonstrate the robustness to the network structures and loss functions respectively. In \Tref{tab:robustness} (Column 2 $\sim$ 5), it shows that, though only UNet based surrogate model trained with $L2$ loss is leveraged in the adversarial training stage, the proposed deep model watermarking framework can resist both white-box and black-box attacks when equipped with the newly proposed deep image-based invisible watermarking technique.
	
	To further demonstrate the robustness to different loss functions, we use the UNet as the default network structure and train surrogate models with different combinations of loss functions. As shown in \Tref{tab:robustness} (Column 6 $\sim$ 10), the proposed deep watermarking framework has a very strong generalization ability and can resist different loss combinations with very high success rates. In \Fref{fig:vis_sr}, we provide one example watermark extracted from different surrogate models.

\subsection{Comparison with other methods} \label{exp:cmp2}

We further compare our method with one representative traditional watermarking algorithm \cite{voloshynovskiy2001multibit}  and one DNN-based watermarking algorithm HiDDeN \cite{zhu2018hidden} by hiding 64-bit as the watermark. For HiDDeN, two variants proposed in their paper are considered: with/without noise layer between the encoder and the decoder. Compared to the ``without noise layer" version, ``with noise layer" version is shown to be more robust. 

As shown in \Tref{tab:cmp2}, our method not only has better embedding and extracting ability, but also has better robustness to surrogate model attack. For the traditional watermarking algorithm \cite{voloshynovskiy2001multibit}, it can only resist the attack of some special surrogate models (only UNet in our experiment) because its extracting algorithms cannot handle the degraded watermarked images from different surrogate models (shown in \Fref{fig:t-sr}). More importantly, they cannot embed high-capacity watermarks like logo images. Although the visual quality is acceptable, the remnants of the original watermark can be discovered from the residual image (\Fref{fig:invisible_wm_qual}), which may be utilized by the attacker. In terms of the embedding time and extracting time, our method is about 0.0051s and 0.007s for a 256$\times$256 image on GPU,  while traditional spatial invisible watermarking algorithm \cite{voloshynovskiy2001multibit} needs 0.0038s and 0.0047s on CPU. We have also tried some other traditional transform domain watermarking algorithms like DCT-based\cite{fang2018screen}, DFT-based\cite{kang2010efficient} and DWT-based\cite{kang2003dwt}, but all of them do not work and achieve $0\%$ success rate. For the DNN-based watermarking algorithm HiDDeN \cite{zhu2018hidden}, no matter whether the noise layer is involved or not, it is totally fragile to surrogate model attacks.

	\begin{figure*}[h]
	\centering
	\includegraphics[width=0.95\linewidth]{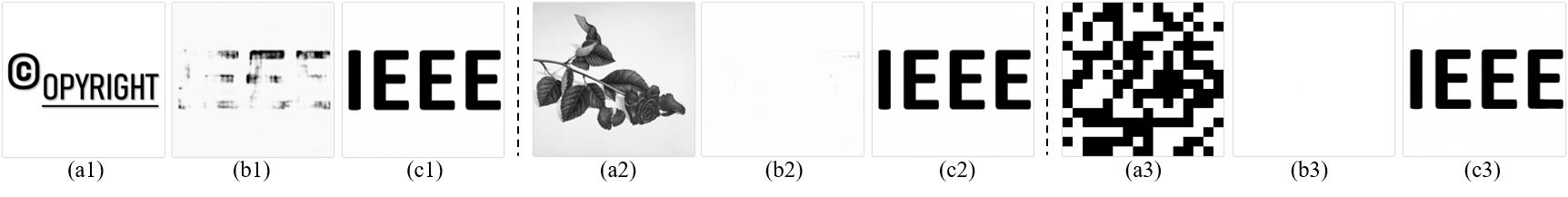}
	\vspace{-1em}
	\caption{Some visual examples of the extractor against watermark overwriting. Three different watermark images (a1 $\sim$ a3) are considered as attacker's watermark, and the corresponding 2nd and 3rd columns are the extracted watermark by our extractor without and with the enhanced training strategy.}
	\label{fig:ow}
	\end{figure*}

\begin{table}[t]
\centering
\caption{Quantitative results with different watermark images for watermark overwriting. The 2nd and 3rd rows are the extracting success rate of our extractor, and $\ddagger$ denotes the results with the enhanced training strategy. We also show the visual quality of surrogate model trained with corresponding re-watermarked images in the last row.}
\vspace{-1em}
\label{tab:ow}
\setlength{\tabcolsep}{1mm}{
\begin{tabular}{c|c|c|c|c}
\hline
Watermarks & None  & ``Copyright"  & ``Flower" & ``BIT"   \\ \hline \hline
$SR_{NC}$$/$$SR_C$       & 100\%$/$100\% & 0\%$/$100\% & 0\%$/$36\% & 0\%$/$0\% \\ \hline
$SR_{NC}$$/$$SR_C$ $\ddagger$     & 100\%$/$100\% & 100\%$/$100\% & 100\%$/$100\% & 99\%$/$100\% \\ \hline
PSNR$/$SSIM        & 25.40$/$0.88 & 25.32$/$0.88 & 25.34$/$0.88 & 25.05$/$0.88 \\ \hline
\end{tabular}}
\end{table}
\subsection{Robustness to Watermark Overwriting} \label{exp:ow}
In real applications, the attackers may follow a similar watermarking strategy and add another watermark upon the model outputs before training the surrogate model, which can potentially destroy the original watermark or cause forensics ambiguity. This type of attack is often called ``watermark overwriting". By using our default setting, we find it cannot resist this attack very well. But following a similar idea as the adversarial training, we find a simple enhanced training strategy can help resist it. Specifically, we first train a single embedding network (simulate the potential overwriting network) to embed ninety diverse watermark images as described in \Sref{mt:ext_mul}, and then add a noise layer between the embedding sub-network and the extractor sub-network in the initial training stage by using the trained embedding network to re-watermark the watermarked images.  We require the extractor to be able to extract the original watermark out from re-watermarked images but blank watermarks from clean images or images only watermarked by the overwriting network. In this way, the extracting ability of the extractor sub-network can be significantly enhanced.

To simulate the attack scenario, we utilize ``IEEE" as the original watermark and use other different watermarks (no overlap with the ninety watermarks) to perform overwriting, which can be regarded as a black-box setting. As shown in \fref{fig:ow} and \Tref{tab:ow},  overwriting with different watermark images makes the extractor fail to some extent, but the extractor sub-network will work well with the above enhanced training strategy. In addition, \Tref{tab:ow} shows a degradation of the performance of surrogate model trained with re-watermarked images, which reveals watermark overwriting is a two-sided sword for the attacker. Need to note that, although our algorithm can still extract our original watermark out, it still suffers from the ambiguity issue (the extractor sub-network from the attacker can also extract their watermark out). And we still need the watermarking protocol to resolve this ambiguity issue.

	\begin{figure*}[t]
		\begin{minipage}[]{0.45\linewidth}
	    \centering
	    \includegraphics[width=0.98\linewidth]{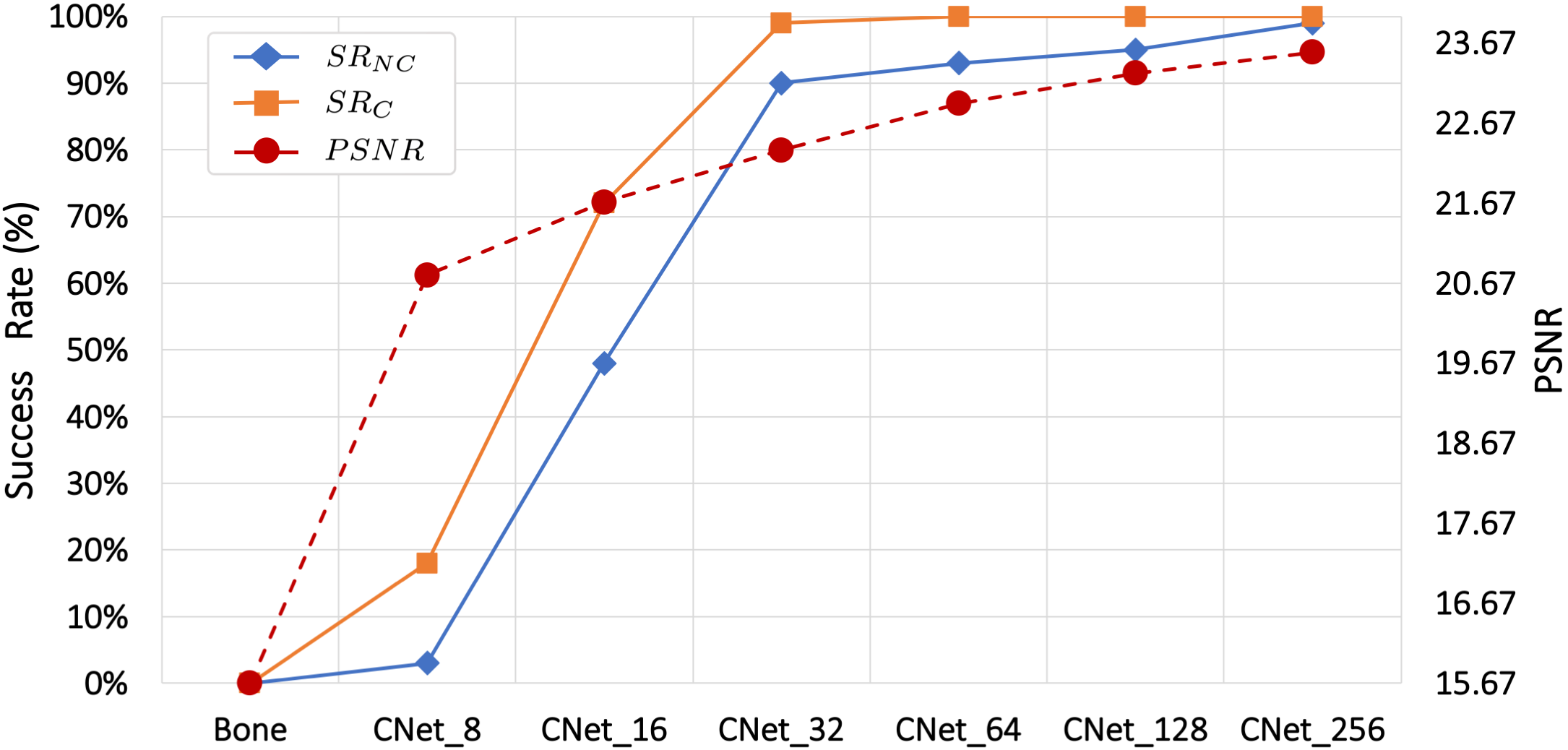}
	\end{minipage}
	\hspace{2mm}
	\begin{minipage}[]{0.5\linewidth}
	    \centering
	    \includegraphics[width=1\linewidth]{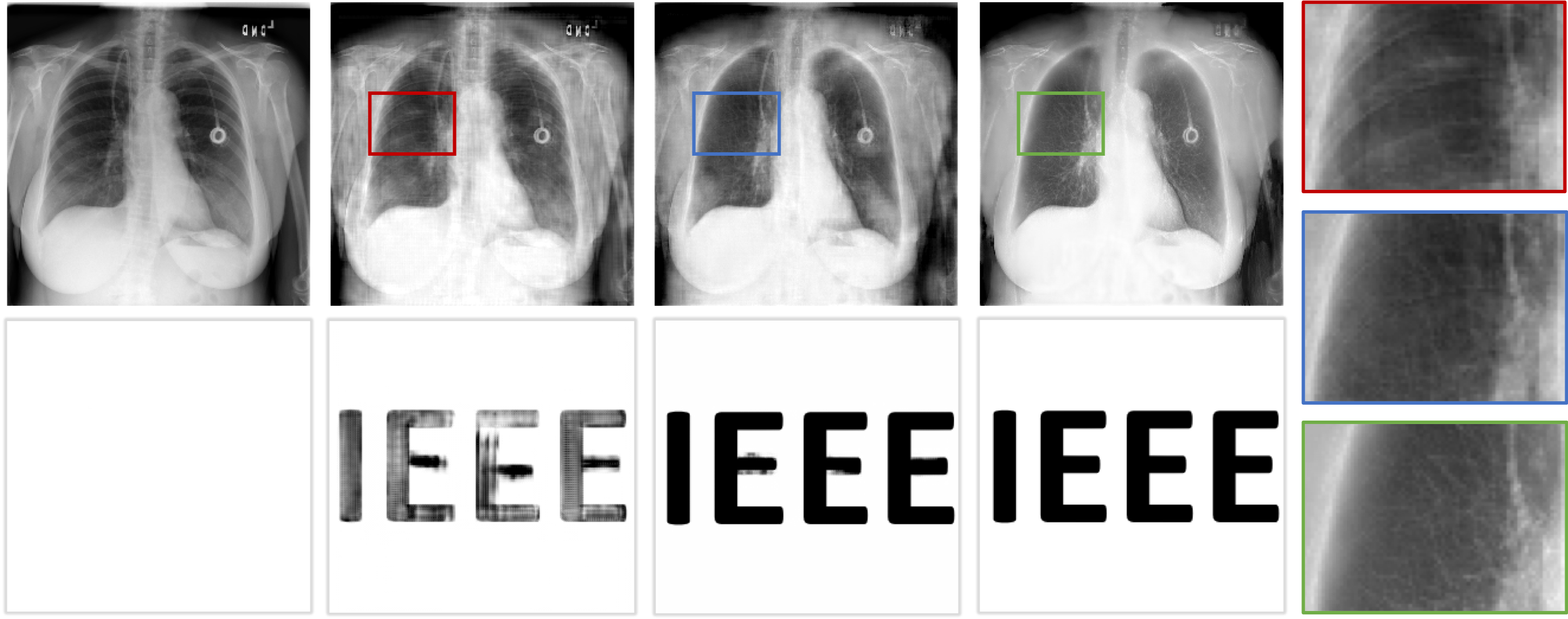}
	\hspace{-6mm}	
	\end{minipage}
	    \caption{The relationship between the extracting ability and the surrogate model performance. \textbf{Left}: the extracting success rates ($SR_{NC},SR_{C}$) and performance (PSNR) change for the surrogate model CNet equipped with different channel numbers (from 8 to 256). 
	     \textbf{Right}: One visual example (the first row): the middle two column represent the surrogate model output for CNet\_8 and CNet\_256 respectively while the first and the last column represent the input image with ``Bone" and the original target model output. The extracted watermarks are displayed in the second row.}
	    \label{fig:channel}
	\end{figure*}

\begin{figure*}[t]
	\centering
	\begin{minipage}[t]{0.47\linewidth}
	\includegraphics[width=\linewidth]{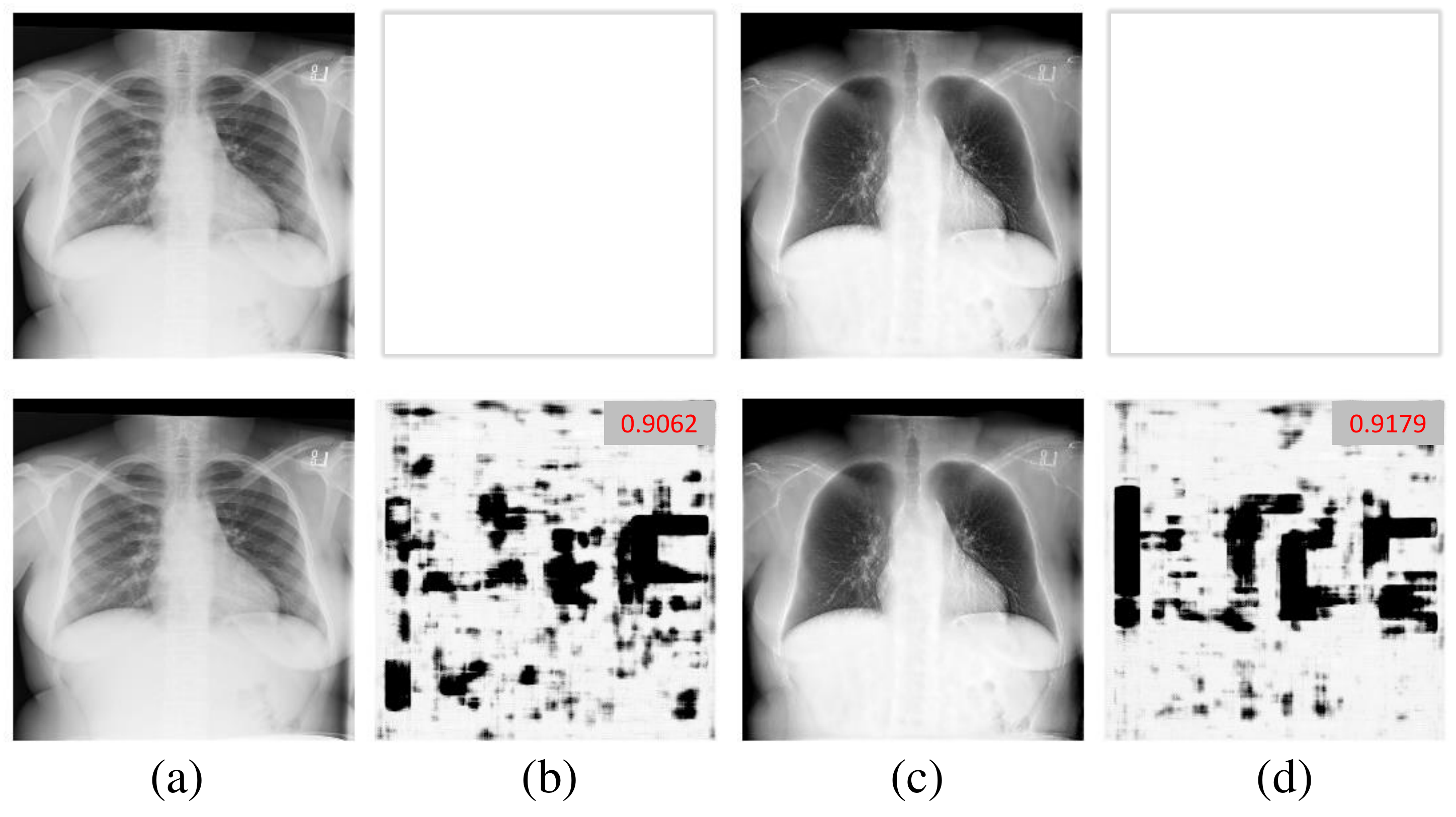}
	\caption{Comparison results with (first row) and without (second row) clean loss: (a) and (c) are the watermark-free images $a_i, b_i$ from domain $\mathbf{A,B}$, (b) and (d)  are  the extracted watermarks  from images $a_i, b_i$ respectively. Number on the topright corner denotes the NC value. } 
	\label{fig:clean_loss} 
	\end{minipage}
	\hspace{4mm}
	\begin{minipage}[t]{0.48\linewidth}
	\centering
	\includegraphics[width=\linewidth]{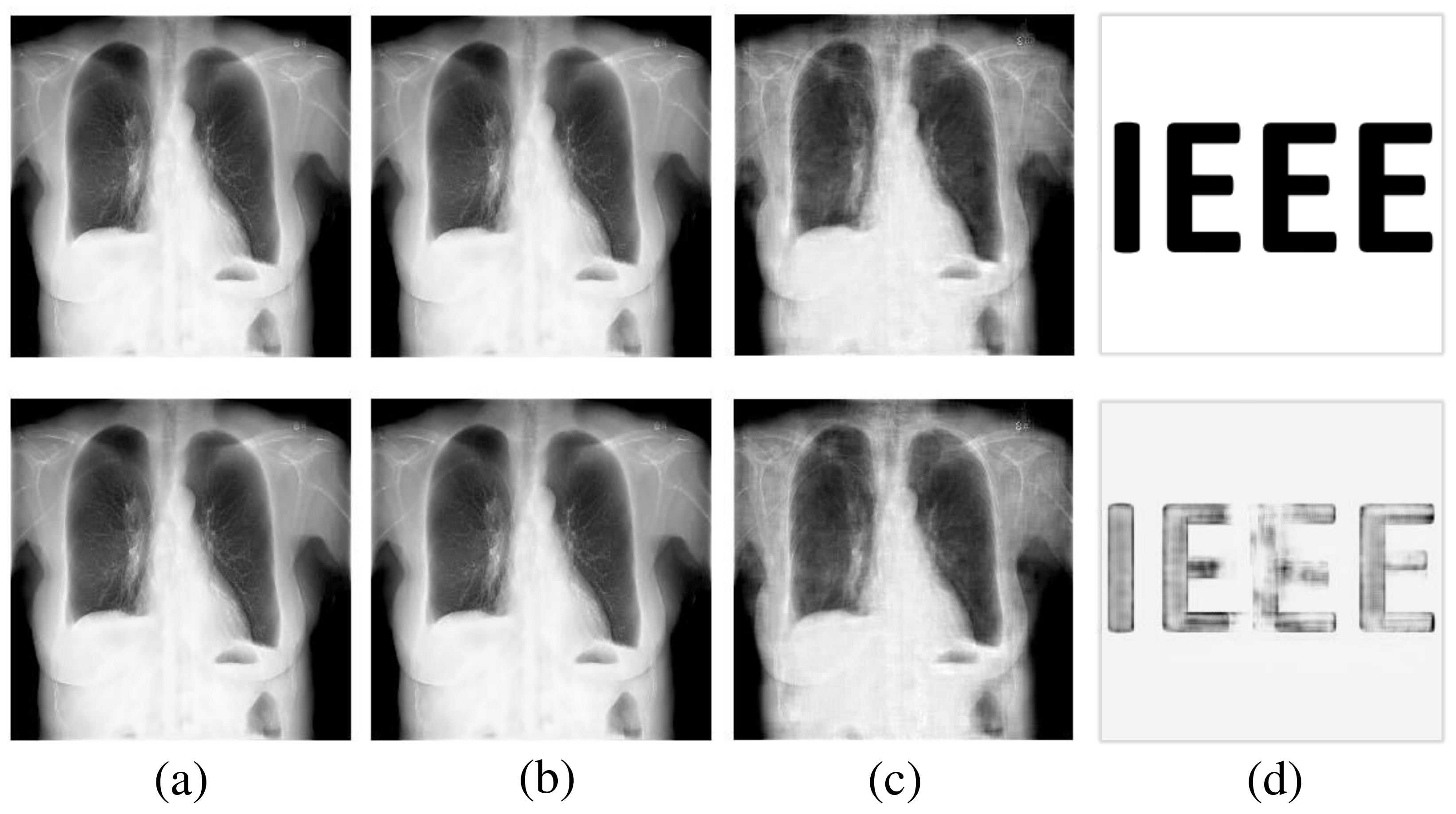}
	\caption{\label{fig:consit_loss} Comparison results with (first row) and without (second row) consistent loss: (a)  watermark-free image $b_i$ from domain $\mathbf{B}$, (b) watermarked image $b_i'$, (c) output $b_i''$ of the surrogate model, (d) extracted watermark out from $b_i''$ .}
	\end{minipage}
\end{figure*}

\subsection{Ablation Study}\label{exp:ab_study}
	\noindent \textbf{Importance of Clean Loss and Consistent Loss.} At the watermark extracting end, besides the watermark reconstruction loss, we also incorporate the extra clean loss for watermark-free images and watermark consistent loss among different watermarked images. To demonstrate their importance, two control experiments are conducted by removing them from the extracting loss. As shown in \Fref{fig:clean_loss}, without clean loss, the extractor will always extract meaningless watermark from watermark-free images of domain $\mathbf{A,B}$ with high NC values, which will make the forensics invalid.
 
	 Similarly in \Fref{fig:consit_loss}, when training without the consistent loss, we find the extractor sub-network $\mathbf{R}$ can only extract very weak watermarks or even cannot extract any watermark out from the outputs of the surrogate mode. This is because the watermark consistency hidden in watermarked image $b_i'$ is destroyed to some extent and makes it more difficult to learn unified watermarks into the surrogate model. By contrast, our method can always extract very clear watermarks out.		

	\vspace{1em}
	\noindent \textbf{Importance of Adversarial Training.} As described above, to enhance the extracting ability of $\mathbf{R}$, another adversarial training stage is used. To demonstrate its necessity, we also conduct the control experiments without adversarial training, and attach the corresponding results in  \Tref{tab:robustness}  (labelled with ``$\dagger$"). It can be seen that, with the default $L$2 loss, its extracting success rate is all about $0\%$ for surrogate models of different network structures. When using UNet as the network structure but trained with different losses, only the embedded watermarks of some special surrogate models can be partially extracted, which demonstrates the significant importance of the adversarial training. In our understanding, the reason why adversarial training with UNet can generalize well to other networks may come from two different aspects: 1) Different surrogate models are trained with the similar task-specific loss functions, so their outputs are similar; 2) As shown in the recent work \cite{wang2020cnn}, different CNN-based image generator models share some common artifacts during the generation process. So training with the degradation brought by UNet makes the extracting network robust to other networks.
	
	\vspace{1em}
    \noindent \textbf{Extracting Ability \textit{vs} Surrogate Model Performance.}
    In this section, we further  analyze the relationship between the extracting ability and the surrogate model performance. Intuitively, the extracting ability will be high when the surrogate model's performance is good, and vice versa. To simulate this case, we take CNet with $L$2 loss as the baseline and adjust its learning ability by changing the feature channel number from 8 to 256. It can be observed from \Fref{fig:channel} that, as the feature channels increase, the performance of surrogate model (PSNR from 20.77 to 23.55) and the extracting ability of extractor $\mathbf{R}$ ( $SR_{NC}$: from 3\% to 99\% and  $SR_{C}$: from 18\% to 100\%) both increase. From the model protection perspective, when the surrogate model does not perform well, we can view it as successful IP protection. For the visual illustration, we give one example for CNet\_8 and CNet\_{256} on the right of \Fref{fig:channel}. It can be seen that our model fails to extract the watermark only when the surrogate model completely fails to conduct the debone task.

\begin{table}[]
\centering
	\caption{\label{tab:lambda} Quantitative results of our method with different size watermark images. We take Debone-IEEE task for example.}
	\vspace{-2mm}
	\setlength{\tabcolsep}{3.3mm}{
\begin{tabular}{c|c|c|c|c}
\hline
$\lambda$& PSNR & SSIM & NC &  $SR_{NC}$  $/$  $SR_C$    \\ \hline
0.1  & 49.82 & 0.9978 & 0.9998 & 100\% $/$ 100\% \\ \hline
0.5  & 48.44 & 0.9969 & 0.9998 &100\% $/$ 100\% \\ \hline
1  & 47.29& 0.9960 & 0.9999 &100\% $/$ 100\% \\ \hline
2 & 45.12 & 0.9934 & 0.9999 &100\% $/$ 100\% \\ \hline
10 & 43.21 & 0.9836  & 0.9999 &100\% $/$ 100\% \\ \hline
\end{tabular}}
\end{table}

\vspace{-2mm}
\begin{table}[]
\centering
	\caption{\label{tab:size} Quantitative results of our method with different size watermark images. We take Debone-IEEE task for example.}
	\label{fg:size_ab}
	\vspace{-2mm}
	\setlength{\tabcolsep}{3.3mm}{
\begin{tabular}{c|c|c|c|c}
\hline
SIZE& PSNR & SSIM & NC &  $SR_{NC}$  $/$  $SR_C$    \\ \hline
32  & 46.89 & 0.9962 & 0.9999 &100\% $/$ 100\% \\ \hline
64  & 47.49 & 0.9965 & 0.9999 &100\% $/$ 100\% \\ \hline
96  & 48.06 & 0.9967 & 0.9999 &100\% $/$ 100\% \\ \hline
128 & 47.68 & 0.9965 & 0.9999 &100\% $/$ 100\% \\ \hline
256 & 47.29 & 0.9960  & 0.9999 &100\% $/$ 100\% \\ \hline
\end{tabular}}
\end{table}

\vspace{1em}
\noindent \textbf{Influence of Hyper-parameter and Watermark Size.} For the hyper-parameter setting, we conduct control experiments with different $\lambda$, which is used to balance the ability of embedding and extracting. As shown in \Tref{tab:lambda}, although the visual quality of watermarked images and the NC value  will be influenced by different $\lambda$, the final results are all pretty good, which means our algorithm is not sensitive to $\lambda$.  We further try different sizes of watermark to test the generalization ability. Since we require the size of watermark to be same as the cover image in our framework, we pad the watermark with 255 if its size is smaller than 256. As shown in \Tref{fg:size_ab}, our method generalizes well to watermarks of different sizes.

	\begin{table}[t]
	\centering
	\caption{Comparison of the original embedding and extracting ability between the multiple-watermark ($\ast$) and per-watermark-per-network setting. Two representative watermarks ``IEEE" and "Flower" are used here.} 
	\setlength{\tabcolsep}{3.5mm}{
    \begin{tabular}{c|c|c|c}
    \hline
        Task & PSNR & SSIM & $SR_{NC}$  $/$  $SR_C$ \\
        \hline
        \hline
         Debone\_IEEE & 47.76 & 0.99 &100\%  $/$ 100\%  \\
         \hline
         Debone\_IEEE $\ast$ & 41.87 & 0.99 & 100\%  $/$ 100\%  \\
         \hline
         Debone\_Flower & 46.36 & 0.99 & 100\%  $/$ 100\% \\
         \hline
         Debone\_Flower $\ast$ & 41.67 & 0.98 & 100\%  $/$ 100\%  \\
         \hline         
    \end{tabular}}
	\label{tab:mu_vis}
	\end{table}

	\begin{figure}
	    \centering
	    \includegraphics[width=0.9\linewidth]{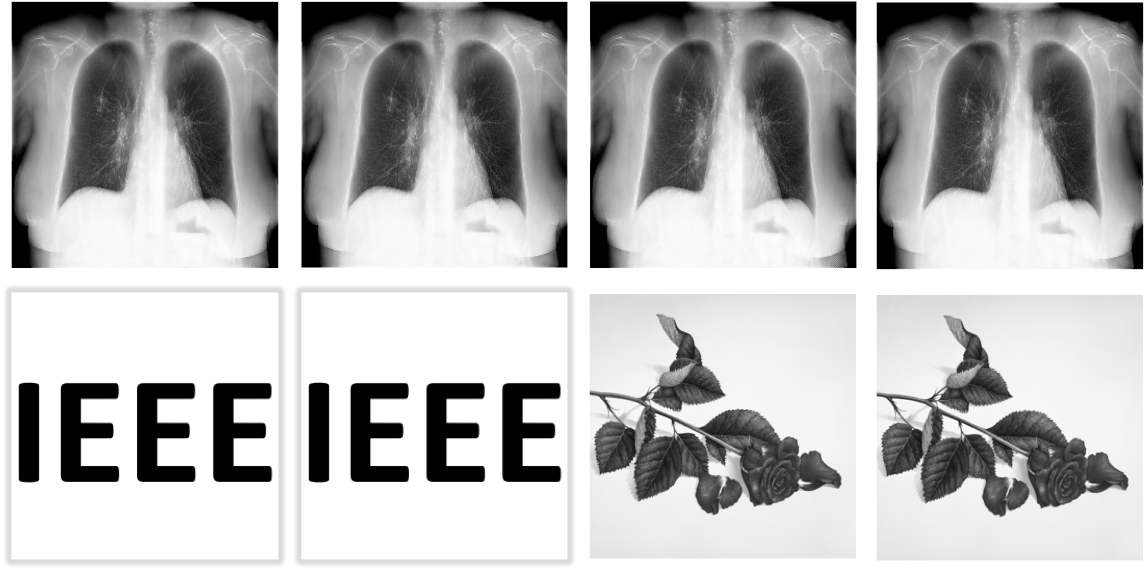}
	    \vspace{-1em}
	    \caption{Visual comparisons between the multiple-watermark (even columns) and the per-watermark-per-network (odd column) setting. The top row are the watermarked images and the bottom row are the corresponding extracted watermarks.}
	    \label{fig:mul_vis} 
	\end{figure}

	\begin{table*}[t]
	    \scriptsize
	    \centering
	    \caption{ Comparison of the success rate ($SR_{NC}$/$SR_{C}$) against surrogate model attack between the multiple-watermark($\ast$) and per-watermark-per-network setting. Column 2 $\sim$ 5 are trained with $L$2 loss but different network structures and Column 6 $\sim$ 10 are trained with UNet network structure but different loss combinations.}	    
	    \setlength{\tabcolsep}{1.3mm}{ 
	    \begin{tabular}{c|c|c|c|c|c|c|c|c|c}
	        \hline
	                     $SR_{NC} / SR_{C}$ & CNet    & Res9 & Res16 &UNet & $L$1    &  $L$2   &  $L$1 + $L_{adv}$     & $L$2 + $L_{adv}$   & $L_{perc}$+$L_{adv}$  \\ 
	         \hline
	         \hline
	                  Debone\_IEEE & 93\% $/$ 100\%    &100\%  $/$ 100\%    & 100\%  $/$ 100\%    &100\%  $/$ 100\%    & 100\% $/$ 100\%    &100\%  $/$ 100\%    & 100\%  $/$ 100\%    &100\%  $/$ 100\%  
	           & 71\% $/$ 94\%   \\
	         \hline
	                   Debone\_IEEE $\ast$ & 89\% $/$ 97\%    &90\%  $/$ 95\%    & 94\%  $/$ 97\%    &100\%  $/$ 100\% & 100\% $/$ 100\%    &100\%  $/$ 100\%    & 100\%  $/$ 100\%    &100\%  $/$ 100\%    & 97\% $/$ 100\%        \\
	            
	         \hline
	                  Debone\_Flower & 73\% $/$ 82\%    &83\%  $/$ 88\%    & 89\%  $/$ 92\%    &100\%  $/$ 100\%  & 100\% $/$ 100\%    &100\%  $/$ 100\%    & 100\%  $/$ 100\%    &99\%  $/$ 100\%    & 100\% $/$ 100\%      \\
	         \hline
	                   Debone\_Flower $\ast$& 94\% $/$ 99\%    &97\%  $/$ 99\%    & 97\%  $/$ 100\%    &100\%  $/$ 100\% & 100\% $/$ 100\%    &100\%  $/$ 100\%    & 100\%  $/$ 100\%    &100\%  $/$ 100\%    & 99\% $/$ 100\%  \\
	         \hline
	    \end{tabular}}
	    \label{tab:mu_net_loss}
	\end{table*}

	\section{Extensions} \label{sec:ext} 
	In this section, we will provide the experiment results for the multiple watermarks and self-watermarked extensions in \Sref{ext:mu} and \Sref{ext:swm} respectively, then discuss how to leverage the proposed framework to protect private data and traditional non-CNN models in \Sref{ext:data}.

    \subsection{Multiple watermarks within one network.} \label{ext:mu} 
    As mentioned in \Sref{mt:ext_mul}, the proposed framework is flexible to embed multiple different watermarks with just a  single embedding and extractor sub-network. To showcase it, we take the debone task for example and select 10 different logo images from the Internet as watermarks for training. For comparison, we use the logo ``IEEE" and ``Flower" as two representatives and compare them with the results of the default per-watermark-per-network setting. In terms of the original embedding and extracting ability, as shown in \Tref{tab:mu_vis}, the visual quality of this multiple-watermark setting measured by PSNR degrades from 47.76 to 41.87 compared to the default setting but is still larger than 40, which is reasonably good. Some visual examples are further displayed in \Fref{fig:mul_vis}. In terms of the robustness to the surrogate model attack, we show the comparison results of different networks and loss functions in \Tref{tab:mu_net_loss}. It can be seen that the extracting success rates of the multiple-watermark setting are comparable to those of the default setting, thus preserving a similar level of robustness.  

	\subsection{Self-Watermarked Model} \label{ext:swm} 

	\begin{table*}[]
	\centering
	\begin{minipage}[t]{0.4\linewidth}
	\centering
	    \caption{The performance comparison between the original target model (``\_original") and self-watermarked target model (``\_Flower").  Obviously, the self-watermarked model can keep the original deraining functionality well while embedding the watermarks in the outputs.}   
    \setlength{\tabcolsep}{2.5mm}{ 
    \begin{tabular}{c|c|c|c}
    \hline
        Task & PSNR & SSIM & $SR_{NC}$  $/$  $SR_C$ \\
        \hline
         \hline
         Derain\_original & 32.49 & 0.93 & NA\\
         \hline
         Derain\_Flower & 32.13 & 0.93 & 100\%   $/$ 100\% \\
         \hline         
    \end{tabular}}
	\label{tab:spe_vis}
	\end{minipage}
	\hspace{5mm}
	\begin{minipage}[t]{0.52\linewidth}
	\centering
	    \caption{Quantitative results of applying the proposed framework to data (DPED \cite{ignatov2017dslr}) and traditional algorithms (RTV \cite{xu2012structure}) protection. The extracting success rate marked with $\ddagger$ denotes the results of resisting an example surrogate model attack ($L$1 + $L_{adv}$). The second to fourth columns represent the results of the learned embedding and extracting sub-network.} 
	\setlength{\tabcolsep}{2.8mm}{   
    \begin{tabular}{c|c|c|c|c}
    \hline
        Task & PSNR & SSIM & $SR_{NC}$  $/$  $SR_C$ & $SR_{NC}$  $/$  $SR_C$ $\ddagger$ \\
        \hline
         \hline
         DPED \cite{ignatov2017dslr} & 46.60 & 0.99 & 100\%   $/$ 100\% & 99\% $/$ 100\% \\
         \hline
         RTV \cite{xu2012structure} & 44.21 & 0.99 & 100\%   $/$ 100\%  &  99\%   $/$ 100\% \\
         \hline         
    \end{tabular}}
    \label{tab:ext3_vis}
	\end{minipage}
	\end{table*}

	\begin{table*}[t]
	    \scriptsize
	    \centering
	    \caption{ The comparison between task-agnostic watermarking model and self-watermarked model  about the success rate ($SR_{NC}$/$SR_{C}$) of resisting the attack from surrogate models trained with  different loss combinations..}	    
	    \setlength{\tabcolsep}{1.3mm}{ 
	    \begin{tabular}{c|c|c|c|c|c|c|c|c|c}
	        \hline
	                     $SR_{NC} / SR_{C}$ & CNet    & Res9 & Res16 &UNet & $L$1    &  $L$2   &  $L$1 + $L_{adv}$     & $L$2 + $L_{adv}$   & $L_{perc}$+$L_{adv}$  \\ 
	         \hline
	         \hline
	                  Task-agnostic &100\%  $/$ 100\%    &100\%  $/$ 100\%    & 100\%  $/$ 100\%    &100\%  $/$ 100\%    & 100\% $/$ 100\%    &100\%  $/$ 100\%    & 100\%  $/$ 100\%    &99\%  $/$ 100\%  
	           & 100\%  $/$ 100\%   \\
	         \hline
	                  Self-watermarked & 100\%  $/$ 100\%    &100\%  $/$ 100\%    & 100\%  $/$ 100\%   &100\%  $/$ 100\% & 100\% $/$ 100\%    &100\%  $/$ 100\%    & 100\%  $/$ 100\%    &100\%  $/$ 100\%    & 100\%  $/$ 100\%         \\
	            
	         \hline
	    \end{tabular}}
	    \label{tab:spe_net_loss}
	\end{table*}

	\begin{figure*}[t]
	    \centering
	    \includegraphics[width=0.9\linewidth]{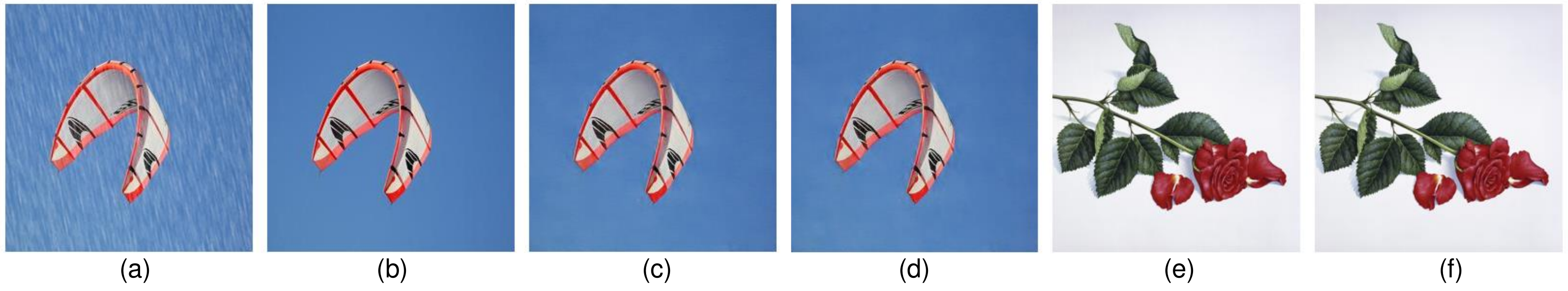}
	    \caption{One visual example of original target model and self-watermarked model: (a) image $a_i$, (b) ground-truth image $(b_0)_i$, (c) output of original target model $b_i$, (d) output of self-watermarked model $b_i^*$, (e) target watermark,  (f) extracted watermark from image $b_i^*$.}
		\label{fig:spe_vis} 
	\end{figure*}

	\begin{figure*}[h! ]
	    \centering
	    \hspace{0.3em}
	    \includegraphics[width=0.9\linewidth]{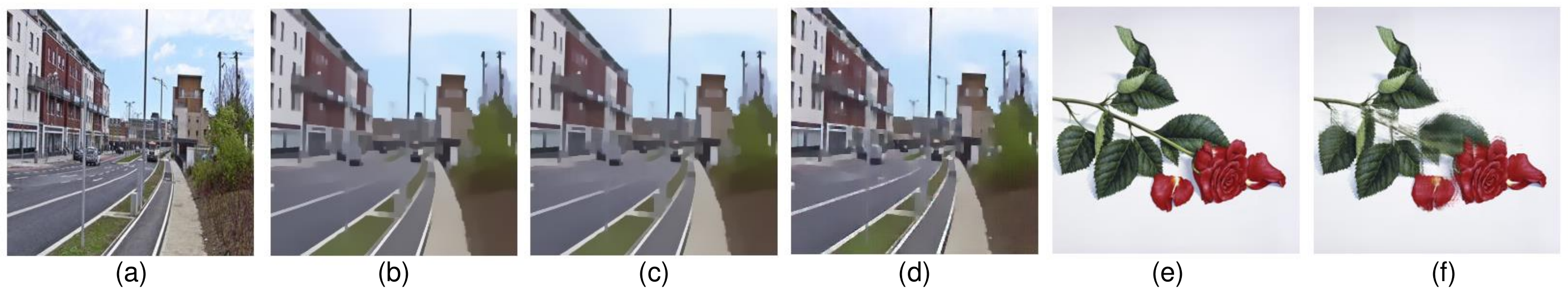}
	    \caption{One visual example of traditional algorithms (RTV \cite{xu2012structure}) protection : (a) image $a_i$, (b) watermark-free image  $b_i$, (c) watermarked image $b_i'$, (d) output of surrogate model (Res9 with  $L$1 + $L_{adv}$), (e) and (f) are the corresponding extracted watermark from (c) and (d).}
		\label{fig:ext3_vis} 
	\end{figure*}
	
	Since the embedding sub-network $\mathbf{H}$ itself is an image processing network and can have different network structures, given a target image processing network $\mathbf{M}$, it is possible to absorb the watermark embedding functionality of $\mathbf{H}$ into $\mathbf{M}$ as shown in \Fref{fig:task_spe_dd}. In this way, $\mathbf{M}$ is self-watermarked without the need of an extra barrier $\mathbf{H}$. To demonstrate this possibility, we use the derain task with the ``Flower" logo image as an example and jointly learn the deraining and watermark embedding process within one single network. 
	In \Tref{tab:spe_vis}, we first compare the deraining performance between the original target model and the self-watermarked target model. It can be seen that the self-watermarked model can achieve very close deraining results (PSNR:32.13 and SSIM: 0.93) to the original target model (PSNR:32.49 and SSIM: 0.93). Two visual examples are further showcased in \Fref{fig:spe_vis}. Besides the deraining functionality, the self-watermarked model can also hide the watermarks well with $0.9999$ NC values. In \Tref{tab:spe_net_loss}, we further demonstrate the robustness to the surrogate model attacks of different network structures and loss functions. It shows that the self-watermarked model is very robust with near $100\%$ extracting success rates.
	
	 \subsection{Protect Valuable Data and Traditional Algorithms} \label{ext:data} 
	 In this paper, though most experiments are conducted to simulate CNN model protection, the proposed idea is very general and can be easily applied to data protection and traditional non-CNN algorithm protection. In details, we can follow the default task-agnostic watermarking setting and embed watermarks into labeled target data or add a watermarking sub-network barrier after traditional algorithms' output.
	 To simulate data protection, we adopt the DPED (``DSLR Photo Enhancement Dataset") dataset \cite{ignatov2017dslr} as an example. It is collected by using high-end cameras and careful alignment to train a high-quality image enhancement network.  For traditional algorithm protection, we consider the famous structure-aware texture smoothing method RTV \cite{xu2012structure} and choose 6000 images from  PASCAL VOC dataset \cite{everingham2010pascal} and COCO dataset \cite{lin2014microsoft} as its inputs. As our default setting, we split the dataset into different parts for embedding sub-network training and surrogate model attack. The detailed evaluation results are given In \Tref{tab:ext3_vis}. On the one hand, the learned embedding sub-network can keep the original visual quality very well with high PSNR/SSIM values and the extracting sub-network has a $100\%$ extracting success rate. On the other hand, we take the Res9 with  $L$1 + $L_{adv}$  as an example surrogate model, and the proposed framework is able to extract the watermark out when the surrogate model is trained with the target dataset or imitates the behavior of the traditional algorithm. Some visual results are shown in \Fref{fig:ext3_vis}.

	\section{Conclusion and Discussion}

	IP protection for deep models is an important but seriously under-researched problem. Inspired by traditional spatial invisible media watermarking and the powerful learning capacity of deep neural networks, we innovatively propose a novel deep spatial watermarking framework for deep model watermarking. To make it robust to different surrogate model attacks and support image-based watermarks, we dedicatedly design the loss functions and training strategy. Experiments demonstrate that our framework can resist the attack from surrogate models trained with different network structures and loss functions. By jointly training the target model and watermark embedding together, we can even make the target model self-watermarked without the need of an extra watermark embedding sub-network. Moreover, though our motivation is to protect deep models, it is general to data protection and traditional algorithms protection. We hope our work can inspire more explorations of deep model IP protection, such as more types of models (detection, semantic segmentation) and protection strategies. 
	
	There are still some interesting questions to explore in the future. For example, though the joint training of the embedding and extracting sub-networks in an adversarial way makes it difficult to find some explicit pattern in the watermarked image, it is worthy to study what implicit watermark is hidden. Besides, our method is not robust enough to some pre-processing techniques for surrogate model attack, such as random cropping and resizing, because the consistency we rely on  will be destroyed. So it is necessary to design some new kind of consistency which is intrinsically robust to such pre-processing.

	{
		\bibliographystyle{IEEEtran}
		\bibliography{references}
	}

\end{document}